\documentclass[12pt]{article}

\usepackage{amsmath,amsthm,amssymb}
\usepackage{xcolor}
\usepackage{algorithm}
\usepackage{algorithmicx}
\usepackage{algpseudocode}
\usepackage{subcaption}
\usepackage{fullpage}

\usepackage{amsmath,amsfonts,amssymb}

\usepackage{multirow}
\usepackage{cellspace}
\usepackage{setspace}
\usepackage{bm}
\usepackage{bbm}
\usepackage{graphicx}
\usepackage{tikz,pgfplots}
\usepackage{framed}
\usepackage[utf8]{inputenc}
\usepackage[T1]{fontenc}
\usepackage[round]{natbib}

\usepackage{hyperref}
\hypersetup{colorlinks,
            linkcolor=blue,
            citecolor=blue,
            urlcolor=magenta,
            linktocpage,
            plainpages=false}

\usepackage[capitalise]{cleveref}

\theoremstyle{plain}
\newtheorem{theorem}{Theorem}
\newtheorem{lemma}[theorem]{Lemma}

\newtheorem*{theorem*}{Theorem}
\newtheorem*{lemma*}{Lemma}
\newtheorem*{corollary*}{Corollary}
\newtheorem*{proposition*}{Proposition}
\newtheorem*{claim*}{Claim}
\newtheorem*{fact*}{Fact}

\theoremstyle{definition}

\newtheorem*{definition*}{Definition}

\newtheorem*{remark*}{Remark}

\newtheorem*{example*}{Example}

\newcommand{\ignore}[1]{}

\DeclareMathOperator*{\argmin}{arg\,min}

\newcommand{\be}{\begin{align}}
\newcommand{\en}{\end{align}}

\newcommand{\ben}{\begin{align*}}
\newcommand{\enn}{\end{align*}}

\newcommand{\reals}{\mathbb{R}}

\def\reals{{\mathcal R}}

\newcommand{\D}{{\mathcal{D}}}

\newcommand{\K}{\mathcal{K}}
\newcommand{\R}{\mathcal{R}}

\def\reals{{\mathbb R}}

\def\bold0{\mathbf{0}}

\def\be{\mathbf{e}}

\newcommand{\B}{\mathcal{B}}

\renewcommand{\O}{O}

\newcommand{\tO}{\wt{\O}}

\newcommand{\regret}{\text{Regret}}
\newcommand{\A}{\mathcal{A}}
\newcommand{\X}{\mathcal{X}}

\newcommand{\tf}{\tilde{f}}

\newcommand{\F}{\mathcal{F}}

\newcommand{\tp}{\tilde{p}}

\newcommand{\tell}{\tilde{\ell}}

\newcommand{\pmin}{p_{\text{min}}}
\newcommand{\sumin}{\sum_{i=1}^n}
\newcommand{\sumtt}{\sum_{t=1}^T}

\newcommand{\expval}[1]{\mathbb{E}\left[ #1 \right]}
\newcommand{\Expect}{\mathbb{E}}

\newcommand{\rA}{\rm{(A)}}
\newcommand{\rB}{\rm{(B)}}

\newcommand{\Var}{\text{Var}}

\newcommand{\uann}[2]{\underset{#2}{\underbrace{#1}}}

\newenvironment{proofarg}[1]{%
  \renewcommand{\proofname}{Proof #1}\proof}{\endproof}

\usepackage{mathtools}
\DeclarePairedDelimiter{\ceil}{\lceil}{\rceil}

\newcommand{\startfoo}{%
    \par\medskip
    \begin{mdframed}[linewidth=1pt]%
    \let\figure\figurehere
    \let\endfigure\endfigurehere
    \ignorespaces
}
\newcommand{\stopfoo}{%
    \unskip
    \end{mdframed}%
    \par\medskip
}

\newcommand{\W}{\mathcal{W}}
\renewcommand{\O}{\mathcal{O}}
\renewcommand{\tO}{\tilde{\mathcal{O}}}

\title{Online Variance Reduction for Stochastic Optimization}

\author{%
Zalán Borsos 
\qquad Andreas Krause 
\qquad Kfir Y. Levy  \\
\small Department of Computer Science, ETH Zurich
}

\date{}

\begin{document}
\maketitle

\begin{abstract}
Modern stochastic optimization methods often rely on uniform sampling which is agnostic to the underlying characteristics of the data.
This might degrade the convergence by  yielding estimates that suffer from a high variance.  
A possible remedy is to employ non-uniform \emph{importance sampling} techniques, which take the structure of the dataset into account. 
In this work, we investigate a recently proposed  setting  which poses variance reduction as an online optimization problem with bandit feedback.
We devise a novel and efficient algorithm for this setting  that finds a sequence of importance sampling distributions competitive with the best fixed distribution in hindsight, the first result of this kind.
While we present our method for sampling datapoints, it naturally extends to selecting coordinates or even blocks of thereof.  Empirical validations underline the benefits of our method in several settings.
\end{abstract}


\section{Introduction}

Empirical risk minimization (ERM) is among the most important paradigms in machine learning, and  is often the strategy of choice due to its generality and statistical efficiency.
In ERM, we draw a set of  samples $\D=\{x_1,\ldots,x_n\}\subset \X$ from the underlying data distribution and we aim to find a solution $w\in\W$ that minimizes the empirical risk, 
\begin{equation} \label{eq:ERM}
  \min_{w\in\W }L(w) := \frac{1}{n}  \sum_{i=1}^n \ell (x_i, w),
\end{equation}
where $\ell: \mathcal{X} \times \W \rightarrow \reals$ is a given loss function, and $\W\subseteq \reals^d$ is usually a compact domain.

In this work we are interested in sequential procedures for minimizing the ERM objective, and relate to such methods as \emph{ERM solvers}.
More concretely, we focus on the regime where the number of samples $n$ is very large,  and it is therefore desirable to employ ERM solvers that only require  few passes over the dataset. There exists a rich arsenal of such efficient solvers which have been investigated throughout the years, with the canonical example from this category being  Stochastic Gradient Descent (SGD).

Typically, such methods  require an unbiased estimate of the loss function at each round, which is usually  generated   by sampling a few points uniformly at random from the dataset.
However, by employing uniform sampling, these methods are insensitive to the intrinsic structure of the data. In case of SGD, for example, some data points might produce large gradients, but they are nevertheless assigned the same probability of being sampled as any other point. This ignorance often results in high-variance estimates, which is likely to degrade the performance.

The above issue can be mended by employing non-uniform importance sampling.
And indeed, we have recently witnessed several  techniques to do so:
\citet{zhao2015stochastic} and similarly \citet{needell2014stochastic}, suggest using prior knowledge on the gradients of each datapoint in order to devise predefined importance sampling distributions.  \citet{NIPS2017_7025} devise adaptive sampling techniques guided by a robust optimization approach. These are only a few examples of a larger body of work 
 \citep{bouchard2015online, alain2015variance, csiba2016importance}.

Interestingly, the recent works of \cite{pmlr-v70-namkoong17a} and \cite{salehi2017} formulate the task of devising importance sampling distributions as an online learning problem with bandit feedback. In this context, they  think of the algorithm, which adaptively chooses the distribution, as a player that competes against the ERM solver. The goal of the player is to minimize the cumulative variance of the resulting (gradient) estimates.  Curiously, both methods rely on some form of the ``linearization trick''\footnote{ By ``linearization trick'' we mean that these methods update according to a first order approximation  of the costs rather than the costs themselves.} 
 to resort to the analysis of the EXP3  \citep{auer2002nonstochastic}.

On the other hand, the theoretical guarantees of the above methods are somewhat limited. Strictly speaking, none of them provides regret guarantees with respect to the best fixed distribution in hindsight:  \citet{pmlr-v70-namkoong17a} only compete with the best distribution among a \emph{subset} of the simplex (around the uniform distribution).  Conversely, \cite{salehi2017} compete against a solution which might perform worse than the best in hindsight up to a multiplicative factor of $3$.

In this work, we adopt the above mentioned online learning formulation, and design novel importance sampling techniques. 
Our adaptive sampling procedure is simple and efficient, and 
in contrast to previous work, we are able to provide regret guarantees with respect to the best fixed point among the simplex.
As our contribution, we
\vspace{-1.5mm}
\begin{itemize}
\setlength\itemsep{0.05em}
\item motivate theoretically why regret minimization is meaningful in this setting, 
\item propose a novel bandit algorithm for variance reduction ensuring regret  of~$\tO(n^{1/3}T^{2/3})$,
\item empirically validate our method and provide an efficient implementation\footnote{The source code is available at  \url{https://github.com/zalanborsos/online-variance-reduction}}.
\end{itemize}
On the technical side, we do not rely on a ``linearization trick'' but rather directly employ a scheme based on the classical 
 Follow-the-Regularized-Leader approach. 
Our analysis entails several technical challenges, most notably handling  unbounded cost functions while only receiving partial (bandit) feedback. Our design and analysis draws inspiration from the seminal works of  \citet{auer2002nonstochastic}  and 
\cite{Abernethy08}. 
Although we present our method for choosing \emph{datapoints}, it naturally applies to choosing \emph{coordinates in coordinate descent} or even \emph{blocks} of thereof \citep{allen2016even,perekrestenko2017faster, nesterov2012efficiency, necoara2011random}.
More broadly, the proposed algorithm can be incorporated in \emph{any sequential algorithm} that relies on an unbiased estimation of the loss. A prominent  application of our method is variance reduction for SGD, which can be achieved by considering  gradient norms as  losses, i.e., replacing $\ell(w,x_i) \leftrightarrow \|\nabla \ell(w,x_i)\|$. With this modification, our method is minimizing the cumulative variance of the gradients throughout the optimization process.
 The latter quantity directly affects the quality of optimization (we elaborate on this in Appendix \ref{appendix:cumulvariance}). 

The paper is organized as follows. In Section \ref{sec:Motivation}, we formalize the online learning setup of variance reduction and motivate why regret is a suitable performance measure. As the first step of our analysis, we investigate the full information setting in Section \ref{sec:full-info}, which serves as a mean for studying the bandit setting in Section \ref{sec:bandit}. Finally, we validate our method empirically and provide the detailed discussion of the results in Section \ref{sec:experiments}.


\section{Motivation and Problem Definition} \label{sec:Motivation}

\begin{figure}[h]
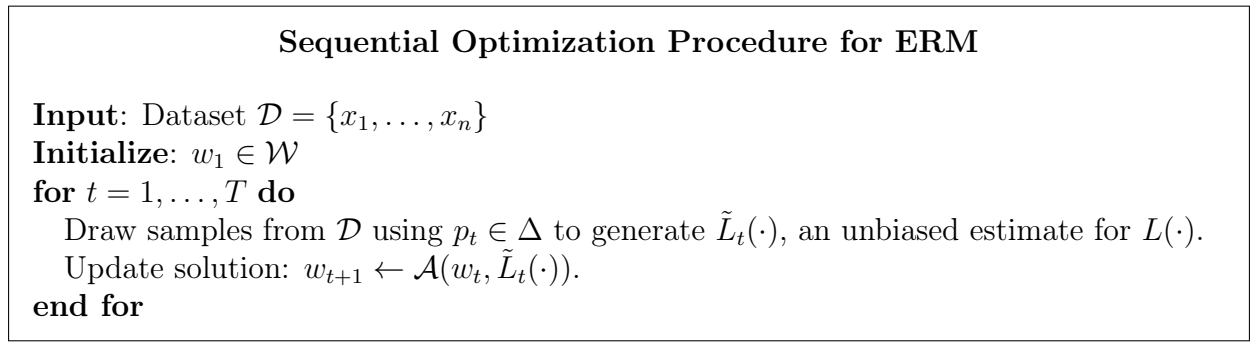

\begin{framed}
\centering{ \textbf{Sequential Optimization Procedure for ERM}\\}
 \flushleft
 \textbf{Input}: Dataset $\D = \{x_1,\ldots,x_n\}$ \\
  \textbf{Initialize}: $w_1 \in \W$ \\
   \textbf{for} $t=1,\ldots, T$ \textbf{do} \\
   \quad {Draw samples from $\D$  using $p_t\in\Delta$ to generate $\tilde{L}_t(\cdot)$,  an unbiased estimate for $L(\cdot)$.} \\
    \quad {Update solution:} $w_{t+1} \gets \A( w_t, \tilde{L}_t(\cdot)$). \\
   \textbf{end for} \\
\end{framed}
\caption{ Template of a sequential  procedure for minimizing the ERM objective.
At each round, we devise a fresh unbiased estimate $\tilde{L}_t(\cdot)$ of the empirical loss, then we update the solution based on the previous solution $w_t$ and $\tilde{L}_t(\cdot)$.}
\label{fig:ERM_Sequential}
\end{figure}

Typical sequential solvers for ERM usually require a fresh unbiased estimate $\tilde{L}_t$  of the loss ${L}_t$ at each round, which is obtained by repeatedly sampling from the dataset.
The template of Figure~\ref{fig:ERM_Sequential} captures a rich family of such solvers such as SGD, SAGA \citep{defazio2014saga}, SVRG \citep{johnson2013accelerating}, and online $k$-Means \citep{bottou1995convergence}.

 A natural way to devise the unbiased estimates $\tilde{L}_t$ is to sample $i_t \in \{ 1, \dots, n\}$ uniformly at random and return $\tilde{L}_t(w) = \ell(x_{i_t},w)$. Indeed, uniform sampling is the common practice when applying SGD, SAGA, SVRG and online $k$-Means. Nevertheless, \emph{any distribution} $p$ in the probability simplex $\Delta$ induces an unbiased estimate.
 Concretely, sampling an index $i \sim p$  induces the estimate
  \begin{equation}
  \tilde{L}(w) := \frac{1}{n \cdot p(i)} \cdot  \ell (x_i, w)
\end{equation}
and it is immediate to show that $\mathbb{E}_{x_i\sim p}[\tilde{L}(w)]=L(w)$.
This work is concerned with efficient ways of choosing a ``good'' sequence of sampling distributions $\{p_1(\cdot),\ldots, p_T(\cdot)\}$.

It is well known that the performance of typical solvers (e.g.  SGD, SAGA, SVRG) improves as the variance of the estimates $\tilde{L}_t(w_t)$ is becoming smaller.
Thus, a natural criterion for measuring the performance of a sampling distribution $p$ is the variance of the induced estimate
\begin{equation*}
\text{Var}_p(\tilde{L}(w)) = \frac{1}{n^2} \sum_{i=1}^n \frac{\ell^2 (x_i, w)}{p(i)} - L^2(w).
\end{equation*}
Denoting $\ell_t(i):=\ell (x_i, w_t)$ and noting that the second term is independent of $p$, we may now cast the task of sequentially choosing the sampling distributions as the  online optimization problem shown in Figure \ref{fig:Protocal_bandit}.
In this protocol, we treat the sequential solver as an adversary that chooses a sequence of loss vectors $\{ \ell_t\}_{t\in[T]}\subset \reals^n$, where $t\in[T]$ denotes $t \in \{ 1, \dots, T\}$. Each loss vector is a function of $w_t$, the solution chosen by the solver in the corresponding round (note that we abstract out this dependence of $\ell_t$ in $w_t$).
The cost\footnote{We use the term ``cost function'' to refer to $f$ in order to distinguish it from the loss $\ell$.} $\frac{1}{n^2}f_t(p_t)$ that the player incurs at  round $t$ is the second moment of the  loss estimate, which is induced by  the distribution chosen by the player at round $t$.
\begin{figure}[h]
\begin{framed}
\centering{ \textbf{Online Variance Reduction Protocol}\\}
 \flushleft
 \textbf{Input}: Dataset $\D = \{x_1,\ldots,x_n\}$ \\
   \textbf{for} $t=1,\ldots, T$ \textbf{do} \\
   \quad{ Player chooses $p_t\in \Delta$.\\
   \quad{ Adversary chooses $\ell_t \in \reals^n$, which induces a cost function $f_t(p):=  \ \sum_{i=1}^n \frac{\ell_t^2(i)}{p(i)}.$} \\
   \quad{ Player draws a sample $I_t\sim p_t$.} }\\
   \quad{ Player incurs a cost $\frac{1}{n^2}f_t(p_t)$, and receives  $\ell_t(I_t)$ as (bandit) feedback.} \\
      \textbf{end for} \\
\end{framed}
\caption{Online variance reduction protocol with bandit feedback}
\vspace{-0.2cm}
\label{fig:Protocal_bandit}
\end{figure}

Next, we define the regret, which is our performance measure for the player,
\begin{equation*}
\regret_T =  \frac{1}{n^2}\left(\sum_{t=1}^Tf_t(p_t) - \min_{p\in \Delta}\sum_{t=1}^T f_t(p) \right).
\end{equation*}
Our goal is to devise a no-regret algorithm such that $\lim_{T \rightarrow \infty} \regret_T / T = 0$, which in turn guarantees that we  recover asymptotically the best fixed sampling distribution. 
In the bandit feedback setting, the player aims to minimize its expected regret $\expval {\regret_T}$, 
where the expectation is taken with respect to the randomized choices of the player and the adversary. Note that we  allow the choices of the adversary to depend on the past choices of the player.

There are few noteworthy comments regarding the above setup. First, it is immediate to verify that the cost functions $f_1, \dots,  f_T$ are convex in $\Delta$, therefore this is an online convex optimization problem. 
Secondly, the cost functions are unbounded in $\Delta$, which poses a challenge in ensuring no-regret.
Finally, notice that the player  receives a \emph{bandit feedback}, i.e., he is allowed to inspect the losses only at the coordinate $I_t$
 chosen at time $t$. To the best of our knowledge, this is the first natural setting
 where, as we will show, it is possible to provide no regret guarantees despite bandit feedback and unbounded costs.

Throughout this work, we assume that the losses are bounded, $l^2_t(i) \leq L$ for all  $i \in [n]$ and \linebreak $t \in [T]$.
Note that our analysis may be extended to the case where the bounds are  instance-dependent, i.e., $l^2_t(i) \leq L_i$ for all  $i \in [n]$ and $t \in [T]$. In practice, it can be beneficial to take into account the different $L_i$'s, as we demonstrate in our experiments.



\subsection{Is Regret  a Meaningful Performance Measure?} \label{sec:Why}
Let us focus on the family of ERM solvers depicted in Figure~\ref{fig:ERM_Sequential}.
As  discussed above, devising loss estimates such that $\tilde{L}_t(w_t)$ has low variance is beneficial for such solvers --- in case of SGD, this is due to strong connection between the cumulative variance of gradients and the quality of optimization that we discuss in more detail in Appendix \ref{appendix:cumulvariance}. 
Translating this observation into the online variance reduction setting  suggests a  natural performance measure:
rather than competing with the best \emph{fixed} distribution in hindsight, we would like to compete against the \emph{sequence} of best distributions per-round  $\left\{p^*_t\gets \argmin\sumin {\ell_t^2(i)}/{p(i)} \right\}_{t\in[T]}$.
This  optimal sequence ensures \emph{zero}  variance  in every round, and is therefore the ideal baseline to compete against.
This also raises the question whether regret guarantees, which  compare against  the best \emph{fixed} distribution in hindsight, are at all meaningful in this context. Note that regret minimization is meaningful in stochastic optimization, when we assume that the losses are generated i.i.d. from some fixed distribution \citep{cesa2004generalization}. Yet, this  certainly does not apply in our  case since  losses are non-stationary and  non-oblivious.

Unfortunately, ensuring guarantees compared to the sequence of best distributions per-round seems generally hard.
However, as we show next, regret  is still a meaningful measure for sequential ERM solvers.
Concretely, recall that our ultimate goal is to minimize the ERM objective. Thus, we are only interested in ERM solvers that actually converge to a (hopefully good) solution for the ERM problem. More formally, let us define $\ell_*(i)$ as follows,
\begin{equation*}
 \ell_{*}(i):=\lim _{t\rightarrow \infty}\ell_t(i),
\end{equation*}
where we recall that $\ell_t(i) :=\ell(x_i,w_t)$, and assume the above  limit to exist for every $i\in[n]$.
We will also denote $L_*:= \frac{1}{n}\sumin  \ell_{*}(i)$.
Moreover, let us assume that the asymptotic solution is better on average than any of the sequential solutions in the following sense,
\begin{equation*}
\frac{1}{T}\sum_{t=1}^T L(w_t) \geq L_*~,\quad \forall T\geq 1
\end{equation*}
where $L(w_t) := \frac{1}{n}\sumin \ell(x_i,w_t)$.
This assumption naturally holds when the ERM solver converges to the optimal solution for the problem, which applies for SGD in the convex case.

The next lemma shows that under these mild assumptions,  competing against the best \emph{fixed}  distribution in hindsight is not  far from competing against the ideal baseline.
\begin{lemma} \label{lem:MeaningfulRegret}
Consider the online variance reduction setting, and for any $i\in[n]$ denote \linebreak
$V_T(i)=\sumtt(\ell_t(i)-\ell_*(i))^2$. Assuming that the losses, $l_t(i)$, are non-negative for all $i \in [n]$, $t\in[T]$, 
 the following holds for any $T\geq 1$,
\begin{align*}
\frac{1}{n^2}\min_{p\in\Delta}\sumtt f_t(p) \leq 
\frac{1}{n^2}\sumtt \min_{p\in\Delta} f_t(p)
~+~
2\sqrt{T}L_* \cdot \frac{1}{n}\sumin  \sqrt{{V_T(i)}} + \left(\frac{1}{n}\sumin  \sqrt{{V_T(i)}} \right)^2~.
\end{align*}
\end{lemma}
Thus, the above lemma connects  the convergence rate of the ERM solver to  the benefit that we get by regret minimization. It shows that the benefit is larger if the ERM solver converges faster.
As an example, let us assume that $|\ell_t(i)-\ell_*(i)| \leq \O(1/\sqrt{t})$, which loosely speaking holds for  SGD.
This assumption implies $V_T(i)\leq \O(\log(T))$, hence by Lemma~\ref{lem:MeaningfulRegret} the regret guarantees translate into guarantees with respect to the ideal baseline, with an additional cost of $\tO(\sqrt{T})$.


\section{Full Information Setting} \label{sec:full-info}
In this section, we analyze variance reduction with full-information feedback.
We henceforth consider the same setting as in Figure \ref{fig:Protocal_bandit}, with the difference that in each round the player receives as a feedback the loss vector at all points $(l_t(1), l_t(2), \dots, l_t(n))$ instead of only $l_t(I_t)$. 
We introduce a new algorithm based on the FTRL approach, and establish an $\O(\sqrt{T})$ regret bound for our method 
in Theorem~\ref{thm:full-info-main}. 
While this setup in itself has little practical relevance, it  
  later  serves as a mean for investigating the bandit setting.

Follow-the-Regularized-Leader (FTRL) is a  powerful approach to online learning problems. According to FTRL, in each round, one selects a point that minimizes the cost functions over past rounds plus a regularization term, i.e., $p_t \gets \argmin_{p\in\Delta} \sum_{\tau=1}^{t-1} f_\tau(p) + \R(p)$. The  regularizer $\R$ usually assures that the choices do not change abruptly over the rounds. We choose \linebreak  $\R(p) =\gamma \sum_{i=1}^n \frac{1}{p(i)}$ which allows to write   FTRL  as   

\begin{equation} \label{eq:ftrl-def}
 p_t \gets \argmin_{p\in\Delta} \sum_{\tau=1}^{t-1}f_\tau(p)  + \gamma \sum_{i=1}^n \frac{1}{p(i)}
~.
\end{equation}
The regularizer $\R(p)=\gamma \sum_{i=1}^n {1}/{p(i)}$ is a natural candidate in our setting, since
it has the same structural form as the cost functions. It also prevents FTRL from assigning vanishing probability mass to any component, thus ensuring that the incurred costs never explode. Moreover, $\R$
assures a closed form  solution to the FTRL as the following lemma shows. 

\begin{lemma} \label{lemma:ftrl-sol}
Denote $l_{1:t}^2(i):=\sum_{\tau=1}^{t}\ell_\tau^2(i)$. The solution to Eq.  \eqref{eq:ftrl-def} is  $p_t(i) \propto \sqrt{\ell_{1:t-1}^2(i)+\gamma}$. 
\end{lemma}
\begin{proofarg}{sketch}
Recalling  $f_t(p) = \sumin \frac{\ell_t^2(i)}{p(i)}$, allows to 
write the FTRL objective  as follows,  
$$\sum_{\tau=1}^{t-1} f_\tau(p)  + \gamma \sum_{i=1}^n {1}/{p(i)}
  = \sumin {(\ell_{1:t-1}^2(i) +\gamma)}/{p(i)}~.$$
It is immediate to validate that the offered solution satisfies the first order optimality conditions in $\Delta$.
Global optimality follows since the FTRL objective is convex in the simplex.

\end{proofarg}

     We are interested in the regret incurred by our method. The following theorem  shows that, despite the non-standard form of the cost functions,  we can obtain $\mathcal{O}(\sqrt{T})$ regret. 
     \begin{theorem}\label{thm:full-info-main} Setting $\gamma=L$, the regret of the FTRL scheme proposed in Equation \eqref{eq:ftrl-def} is: 
\begin{equation*}
\regret_T \leq \frac{27\sqrt{L}}{n}\left(\sum_{i=1}^n \sqrt{ \ell_{1:T}^2(i) }  \right) +44L.
\end{equation*}
Furthermore, since $\ell^2_t(i) \leq L$ we have
$\regret_T \leq 27L\sqrt{T} +44L$.
\end{theorem}

    Before presenting the proof, we briefly describe it. Trying to apply the classical FTRL regret bounds, we encounter a difficulty, namely that the regularizer in Equation \eqref{eq:ftrl-def} can be unbounded. To overcome this issue, we first consider competing with the optimal distribution on a restricted simplex  where $\R(\cdot)$ is bounded. Then we investigate  the cost of considering  the restricted simplex instead of the full simplex. 
 
 Along the lines described above,  
consider the simplex $\Delta$ and the restricted simplex \linebreak $\Delta'=\{ p\in \Delta |  \; p(i) \geq \pmin, \forall i \in [n] \}$ where  $\pmin \leq 1/n$ is to be defined later. We can now decompose the regret  as follows,
\begin{align} \label{eq:regret}
n^2 \cdot \regret_T
=
 \underset{\rA}{\underbrace{ \sumtt f_t(p_t)  - \min_{p \in \Delta'} \sumtt f_t(p)   }} 
 +  
 \underset{\rB}{\underbrace{\min_{p \in \Delta'} \sumtt f_t(p) - \min_{p \in \Delta} \sumtt f_t(p)   }}.
\end{align}
We continue by separately bounding the above terms.
To bound $\rA$, we will use standard tools which relate the regret to the stability of the FTRL decision sequence (FTL-BTL lemma). Term $\rB$ is  bounded by a direct calculation of the minimal values in $\Delta$ and $\Delta'$.

The following lemma bounds term $\rA$.
\begin{lemma} \label{lemma:ub-a} Setting $\gamma =L$, we have:
\begin{equation*}
\sumtt f_t(p_t)   - \min_{p \in \Delta'} \sumtt f_t(p)  \leq 
22n\sqrt{L} \cdot \left( \sum_{i=1}^n \sqrt{ \ell_{1:T}^2(i) } \right) +22n^2L+\frac{nL}{\pmin}.
\end{equation*}
\end{lemma}
\begin{proofarg}{sketch of Lemma  \ref{lemma:ub-a} } 
The regret of FTRL may be related to the stability of the online decision sequence as shown in
 the following lemma due to \cite{kalai2005efficient} (proof 
 can also be found in \cite{Hazan09}  or in \cite{shalev2012online}):
\begin{lemma} \label{Lemma:FTL-BTL}
Let $\K$ be a convex set and  $\R:\K\mapsto\reals$ be a regularizer. Given a sequence of cost functions $\{f_t \}_{t\in[T]}$ defined over $\K$, then setting $p_t = \argmin_{p\in\Delta}\sum_{\tau=1}^{t-1}f_{\tau}(p)+\R(p)$ 
ensures,
\begin{align*}
\sum_{t=1}^T f_t(p_t)-\sum_{t=1}^T f_t(p)\leq \sum_{t=1}^T\left( f_t(p_t) -f_t(p_{t+1}) \right) +(\R(p)-\R(p_1)), \quad \forall p\in\K
\end{align*}
\end{lemma}

Notice that $\R(p) =  L \sum_{i=1}^n {1}/{p(i)}$ is non-negative and bounded by $nL/\pmin$ over $\Delta'$. Thus,  applying the above lemma implies that $\forall \; p\in\Delta'$,
\begin{equation*}
\sumtt f_t(p_t)   - \sumtt f_t(p)  \leq  \sumtt \left( f_t(p_t)- f_t(p_{t+1}) \right) +  \frac{nL}{\pmin}  \leq \sum_{t=1}^T \sumin \ell_t^2(i) \left( \frac{1}{p_t(i)}  - \frac{1}{p_{t+1}(i)}\right) + \frac{nL}{\pmin}~.
\end{equation*}
Using the closed form solution for the  $p_t$'s (see Lemma.~\ref{lemma:ftrl-sol}) enables us to upper bound the last term as follows,
\begin{align} \label{eq:frtl-ub0}
\sum_{t=1}^T \sum_{i=1}^n \ell_t^2(i) \left( \frac{1}{p_t(i)}  - \frac{1}{p_{t+1}(i)}\right)
& \leq 
22n\sqrt{L} \sumin \sqrt{ \ell_{1:T}^2(i) +L}~.
\end{align}
Combining the above with $\sqrt{a+b}\leq \sqrt{a}+\sqrt{b}$ completes the proof.

\end{proofarg}
The next lemma bounds term $\rB$.
\begin{lemma}\label{lemma:ub-b}
\begin{equation*}
\min_{p \in \Delta'} \sumtt f_t(p)  - \min_{p \in \Delta} \ \sumtt f_t(p) \leq 6 n \cdot \pmin \cdot \left( \sumin \sqrt{  \ell_{1:T}^2(i)} \right)^2
\end{equation*}
\end{lemma}
\begin{proofarg}{sketch of Lemma~\ref{lemma:ub-b}}
Using first order optimality conditions we are able show that the  minimal value of the $\sumtt f_t(p)$ over $\Delta$ is exactly $\left(\sumin\sqrt{\ell_{1:t}^2(i)} \right)^2$.
Similar analysis allows to extract a closed form solution to the best in hindsight over $\Delta'$. This in turn enables to upper bound the minimal value over $\Delta'$ by
$\left(\sumin\sqrt{\ell_{1:t}^2(i)} \right)^2/\left(1- n \cdot \pmin\right)^2$. Combining these bounds together with $\pmin\leq 1/2n$ we are able to prove the lemma.

\end{proofarg}

\begin{proofarg}{of Theorem \ref{thm:full-info-main}}
Combining Lemma \ref{lemma:ub-a} and \ref{lemma:ub-b}, we have after dividing by $n^2$,
\begin{equation*}
\regret_T \leq \frac{22\sqrt{L}}{n} \cdot \left( \sum_{i=1}^n \sqrt{ \ell_{1:T}^2(i) } \right) +22L+\frac{L}{n \cdot \pmin}+ \frac{6 \cdot \pmin}{n} \cdot \left( \sumin \sqrt{  \ell_{1:T}^2(i)} \right)^2
\end{equation*}
Since the choice of $\pmin$ is arbitrary and is relevant only for the theoretical analysis, we can set it to $\pmin = \min \left\{1/(2n), \, \sqrt{L}/\left( \sqrt{6}  \sumin \sqrt{  \ell_{1:T}^2(i)} \right) \right\}$ that yields the final result.
\end{proofarg}

\newpage
\section{The Bandit Setting} \label{sec:bandit}
In this section, we investigate the bandit setting (see Figure~\ref{fig:Protocal_bandit}) which is of great practical appeal as we described in Section~\ref{sec:Motivation}.
Our method for the bandit setting is depicted in Algorithm~\ref{alg:bandit}, and it ensures a bound of $\tilde{O}(n^{1/3}T^{2/3})$ on the expected  regret (see Theorem~\ref{thm:bandit-non-oblivious}). Importantly, this bound holds even for non-oblivious adversaries.
The design and analysis of our method builds on some of the ideas that appeared in the  seminal work of  \cite{auer2002nonstochastic}. 

Algorithm~\ref{alg:bandit} is using the bandit feedback in order to design an unbiased estimate of the true loss  $(\ell_t(1),\ldots,\ell_t(n))$ in each round. These estimates are then used instead of the true losses by the full information FTRL algorithm that was analyzed in the previous section.  We do not directly play according to the FTRL predictions but rather mix them with a uniform distribution. Mixing is necessary in order to ensure that the   loss estimates are bounded, which is  a crucial  condition used in the analysis. Next we elaborate  on our method and its analysis.

The algorithm samples\footnote{The
sampling and update in the presented form have a complexity of $\mathcal{O}(n)$.
There is a standard way to improve this based on segment trees that gives $\mathcal{O}(\log n)$ for sampling and update. A detailed description of this idea can be found in section A.4. of \cite{salehi2017stochastic}. The efficient implementation of the sampler is available at \url{https://github.com/zalanborsos/online-variance-reduction} }  an arm $I_t\sim \tp_t$ at every round and receives a bandit feedback $\ell_t(I_t)$. 
This may be used in order to construct an estimate of the true (squared) loss as follows,

\begin{equation*}
\tell_t^2(i) := \frac{\ell_t^2(i)}{\tp_t(i)}\cdot \mathbbm{1} _{I_t=i}~,
\end{equation*}
and it is immediate to validate that the above is unbiased in the following sense, 
\begin{equation*}
\mathbb{E}[\tell_t^2(i) \vert \tp_t,\ell_t] = \ell_t^2(i),\quad \forall i\in[n].
\end{equation*}
Analogously to the previous section it is natural to define modified cost functions as 
\begin{equation*}
\tf_t(p) = \sumin {\tell_t^2(i)}/{p(i)}~.
\end{equation*}
Clearly, $\tf_t$ is an unbiased estimate of the true cost, $\mathbb{E}[\tf_t(p) \vert \tp_t,\ell_t] = f_t(p)$.
From now on  we omit the conditioning on $\tp_t,\ell_t$ for notational brevity. 

Having devised an unbiased estimate, we could return to the full information analysis of FTRL with the modified losses. However, this poses a difficulty, since the modified losses  can possibly be unbounded. We remedy this by mixing the FTRL output, $p_t$, with a uniform distribution.
Mixing encourages exploration, and in turn gives a handle on the possibly unbounded modified losses. Let $\theta\in [0,1]$, and define
\begin{equation*}
\tp_t(i) = (1-\theta)\cdot p_t(i) +{\theta}/{n}.
\end{equation*}
Indeed, since $\tp_t(i) \geq \theta / n$, we have $\tell_t^2(i) \leq nL/\theta$. 
  \begin{algorithm}[t]
            \caption{Variance Reducer Bandit (VRB)}
            \label{alg:bandit}
            \begin{algorithmic}
     \State {\bfseries Input:} $\theta$, $L$, $n$
     \State Initialize $w(i)=0$ for all $i \in [n].$
     \For{$t=1$ {\bfseries to}  $T$}
       \State $p_t(i) \propto \sqrt{w(i) + L\cdot n / \theta}$
      \State $\tp_t(i) = (1-\theta) \cdot p_t(i) + {\theta}/{n}$,  for all $i \in [n]$
        \State Draw $I_t \sim \tp_t$ and play $I_t$.
        \State Receive feedback $l_t(I_t)$, and update $w(I_t) \gets w(I_t) + l^2_t(I_t) / \tp_t(I_t)$.
     \EndFor
  \end{algorithmic}
     \end{algorithm}

We start with analyzing the \emph{pseudo-regret} of our algorithm, where we compare the cost incurred by the algorithm to the cost incurred by the optimal distribution \emph{in expectation}. The pseudo-regret is defined below,
\begin{equation} \label{eq:bandit-regret}
\frac{1}{n^2}\min_{p \in \Delta }  \expval{   \sumtt f_t(\tp_t)  -  \sumtt f_t(p)},
\end{equation}
where the expectation is taken with respect to both the player's choices and the loss realizations. The pseudo-regret is only a lower bound for the \emph{expected regret}, with  an equality when the adversary is oblivious, i.e., does not take the past choices of the player into account.

\begin{theorem} \label{thm:bandit-main}
Let $\theta = (n/T)^{1/3}$.
Assuming $T \geq n$, Algorithm~\ref{alg:bandit} ensures the following bound,
\begin{equation*}
\frac{1}{n^2}\min_{p \in \Delta }  \expval{   \sumtt f_t(\tp_t)  -  \sumtt f_t(p)} \leq 74Ln^{\frac{1}{3}}T^{\frac{2}{3}}.
\end{equation*}
\end{theorem}

\begin{proofarg}{sketch of Theorem~\ref{thm:bandit-main}}
Using the unbiasedness of the modified costs we have
\begin{equation*}
 \min_{p \in \Delta } \expval{   \sumtt f_t(\tp_t)  - \sumtt f_t(p)} 
= \min_{p \in \Delta }  \expval{   \sumtt \tf_t(\tp_t)  -  \sumtt \tf_t(p)} .
\end{equation*}
We can decompose $\frac{1}{n^2} \min_{p \in \Delta } \expval{   \sumtt \tf_t(\tp_t)  -  \sumtt \tf_t(p)} $  into the following terms:
\begin{align*}
&\uann{ \frac{1}{n^2}\expval {\sumtt \tf_t(\tp_t) - \sumtt \tf_t(p_t)  } } {\rA} + \uann{\frac{1}{n^2} \min_{p \in \Delta } \expval{\sumtt \tf_t(p_t)  - \sumtt \tf_t(p)   } }{\rB}
\end{align*}
where $\rA$ is the cost we incur by mixing, and $\rB$ is upper bounded by the regret of playing FTRL with the modified losses. Now we inspect each term separately. 

An upper bound of $  \theta  L  T$ on $\rA$ results from the following simple observation:
\begin{equation*}
\frac{1}{\tp_t(i)} - \frac{1}{p_t(i)} \leq n \theta.
\end{equation*}
For bounding $\rB$, notice that  $p_t$ is   performing FTRL over the modified cost sequence. Combining this together the  bound $\tell_t^2(i) \leq nL/\theta$ allows us to apply Theorem \ref{thm:full-info-main} and get,
\begin{equation}\label{eq:RegretModifiedCosts}
\frac{1}{n^2}\left( \sumtt \tf_t(p_t)  - \min_{p \in \Delta } \sumtt \tf_t(p) \right)
 \leq 
  27 \sqrt{\frac{L}{n \theta}}\left(\sum_{i=1}^n \sqrt{ \tell_{1:T}^2(i) }  \right) +\frac{44nL}{\theta} ~.
\end{equation}
Due to Jensen's inequality we have
\begin{equation*}
\expval{\sum_{i=1}^n  \sqrt{\tell_{1:T}^2(i)}} 
 \leq \sum_{i=1}^n \sqrt{\expval{\tell_{1:T}^2(i)}} = \sum_{i=1}^n \sqrt{\ell_{1:T}^2(i)} ~.
\end{equation*}
Putting these results together, we get an upper bound on the pseudo-regret which we can optimize in terms of $\theta$:
\begin{equation*}
\frac{1}{n^2}  \min_{p \in \Delta }  \expval{   \sumtt f_t(\tp_t)  -  \sumtt f_t(p)} \leq \theta LT +27\sqrt{\frac{L}{n \theta}}\left(\sum_{i=1}^n \sqrt{ \ell_{1:T}^2(i) }  \right) +\frac{44nL}{\theta} .
\end{equation*}
Using the bound $\sum_{i=1}^n \sqrt{ \ell_{1:T}^2(i) }\leq n \sqrt{LT}$ and since we assumed $T\geq n$, we can set \linebreak $\theta = (n/T)^{1/3}$ to get the result. Note that  $\theta$ is dependent on knowing $T$ in advance. If we do not assume that this is possible, we can use the ``doubling trick'' starting from $T=n$ and incur an additional constant multiplier in the regret.
\end{proofarg}

Ultimately, we are interested in the \emph{expected regret}, where we allow the adversary to make decisions by taking into account the player's past choices, i.e., to be \emph{non-oblivious}. 
Next we present the main result of this paper, which establishes a $\tO(n^{1/3}T^{2/3})$  regret bound, where the $\tO$ notation hides the logarithmic factors.
\begin{theorem} \label{thm:bandit-non-oblivious}
Assuming $T \geq n$, the following holds for the expected regret,
\begin{equation*}
\frac{1}{n^2} \expval{   \sumtt f_t(\tp_t)  - \min_{p \in \Delta }  \sumtt f_t(p)} \leq \tilde{\mathcal{O}}\left(Ln^{\frac{1}{3}}T^{\frac{2}{3}}\right).
\end{equation*}
\end{theorem}

\begin{proofarg}{sketch of Theorem~\ref{thm:bandit-non-oblivious}}
Using the unbiasedness of the modified costs allows to decompose the regret as follows,
\begin{align} \label{eq:MasterNonOblivious} 
n^2\expval{ \regret_T } 
&=  
\expval{   \sumtt f_t(\tp_t)  - \min_{p \in \Delta }  \sumtt f_t(p)}  \nonumber\\
&=
\expval{   \sumtt \tf_t(\tp_t)  - \min_{p \in \Delta }  \sumtt \tf_t(p)}
+\expval{  \min_{p \in \Delta }  \sumtt \tf_t(p)  - \min_{p \in \Delta }  \sumtt f_t(p)}  \nonumber\\
&\leq
n^2\mathcal{O}(Ln^{1/3}T^{2/3}) 
+ 
\expval{  
\underset{\rA}{ \underbrace{
 \left(\sumin \sqrt{\tell_{1:T}^2(i)}  \right)^2 -  \left(\sumin \sqrt{\ell_{1:T}^2(i)}  \right)^2
 }}
 },
\end{align}
where  the last line uses  Equation~\eqref{eq:RegretModifiedCosts}  together with  Jensen's inequality (similarly to the proof of Theorem~\ref{thm:bandit-main}). We have also used the closed form solution for the minimal values of  $\sum_t f_t(p)$ and $\sum_t \tf_t(p)$ over the simplex. 

Our approach  to bounding  the remaining term is to establish  high probability bound for $\rA$. In order to do so we shall bound the following differences $\tell_{1:T}^2(i) - \ell_{1:T}^2(i)$. This can be done by applying the 
\vspace{10pt}
appropriate concentration results described below. \\
\textbf{Bounding  $\tell_{1:T}^2(i) - \ell_{1:T}^2(i)$.}
Fix $i\in[n]$ and define $Z_{t,i}:=\tell_t^2(i) - \ell_t^2(i)$. Recalling that \linebreak $\mathbb{E}[\tell_t^2(i) \vert \tp_t,\ell_t] = \ell_t^2(i)$, we have that $\{ Z_{t,i}\}_{t\in[T]}$ is a martingale difference sequence with respect to the filtration $\{\F_{t}\}_{t\in[T]}$ associated with the history of the strategy.  This allows us to apply a version of Freedman's inequality \citep{freedman1975tail}, which bounds the 
sum of differences with respect to their cumulative conditional variance.
Loosely speaking, Freedman's inequality implies that w.p. $\geq 1-\delta$,
$$
\tell_{1:T}^2(i) - \ell_{1:T}^2(i) \leq \tO\left(  \sqrt{ \sum_{t=1}^T \Var(Z_{t,i} \vert \F_{t-1})} \right).
$$
Importantly, the sum of conditional variances can be related to the regret.
Indeed let $p^*$ be the best distribution in hindsight, i.e., $p^* = \arg\min \sumtt f_t(p)$, and define 
$$
 n^2\regret_T(i) = \sumtt \frac{\ell_t^2(i)}{\tp_t(i)} - \sumtt  \frac{\ell_t^2(i)}{p^*(i)}$$
 Then the following can be  shown,
 $$
 \sumtt \Var(Z_{t,i} \vert \F_{t-1}) 
 = \tO\left(n^2 L \cdot \regret_T(i) +    \frac{\ell_{1:T}^2(i)}{p^*(i)} \right).
 $$
To simplify the proof sketch, ignore the second term. Plugging this back into Freedman's inequality  we get,
\begin{equation} \label{eq:BoundReg_i}
\tell_{1:T}^2(i) - \ell_{1:T}^2(i) \leq \tO\left(   \sqrt{ n^2 L \cdot  
\regret_T(i)
} \right).
\end{equation}
\textbf{Final  bound.} Combining the above with the definition of $\rA$
one can to show that w.p. $\geq 1-\delta$,
\begin{equation*}
{\rA} \leq \tO \left(  n\sqrt{LT}\sumin \left( n^2L \cdot \regret_T(i) \right)^\frac{1}{4} \right).
\end{equation*}
Since $\rA$ is bounded by $\text{poly}(n,T)$, we can take 
a small enough $\delta = 1/\text{poly}(n,T)$ such that,
\begin{align*}
\expval{\rA} 
&\leq 
\tO \left(  n^{3/2}L^{3/4}T^{1/2} \cdot
\expval{\sumin 
\left( \regret_T(i) \right)^{1/4} }
 \right)\\
&\leq 
\tO \left(   n^{3/2}L^{3/4}T^{1/2} \cdot
\sumin 
\left( \expval{\regret_T(i) }\right)^{1/4}
 \right)\\
 &\leq 
\tO \left(  n^{9/4}L^{3/4}T^{1/2} \cdot
\left( \expval{ \regret_T}  \right)^{1/4}
 \right)
\end{align*}
where the second line uses Jensen's inequality with respect to the concave function \linebreak $h(u) = u^{1/4}$, and the last line uses $\sumin \regret_T(i) = \regret_T$ together with the fact \linebreak that $\sumin x_i^{1/4} \leq n^{3/4}\left(\sumin x_i\right)^{1/4}$, which is also a consequence of Jensen's inequality since 
$\frac{1}{n}\sumin x_i^{1/4}\leq \left(\frac{1}{n}\sumin\right)^{1/4}$. 
Plugging the above bound back into Eq.~\eqref{eq:MasterNonOblivious} we are able to establish the proof. The full proof is deferred to Appendix~\ref{sec:ProofsExpectedRegret}.
Note that in the full proof we do not explicitly relate the conditional variances to the regret, but this is rather more implicit in the analysis.
\end{proofarg}

\section{Experiments} \label{sec:experiments}
\subsection{Image Classification}

Training a binary classifier with imbalanced data is a challenging task in machine learning. Practices for dealing with imbalance include optimizing class weight hyperparameters, hard negative mining \citep{shrivastava2016training} and synthetic minority oversampling \citep{chawla2002smote}. 
Without accounting for imbalance, the minority samples are often misclassified in early stages of the iterative training procedures, resulting in high loss and high gradient norms associated with these points. Importance sampling schemes for reducing the variance of the gradient norms will sample these instances more often at the early phases, offering a way of tackling imbalance.

For verifying this intuition, we perform the image classification experiment of \cite{bouchard2015online}. We train one-vs-all logistic regression Pascal VOC 2007 dataset \citep{everingham2010pascal} with image features extracted from the last layer of the VGG16 \citep{simonyan2014very} pretrained on Imagenet. We measure the average precision by reporting its mean over the 20 classes of the test data. The optimization is performed with AdaGrad \citep{duchi2011adaptive}, where the learning rate is initialized to 0.1. The losses received by the bandit methods are the norms of the logistic loss gradient.
We compare our method, Variance Reducer Bandit (VRB), to:
\vspace{-1.5mm}
\begin{itemize}
\setlength\itemsep{0.05em}
\item uniform sampling for SGD,
\item Adaptive Weighted SGD (AW) \citep{bouchard2015online} ---  variance reduction by sampling from a chosen distribution whose parameters are optimized alternatingly with the model parameters,
\item MABS \citep{salehi2017} --- bandit algorithm for variance reduction that relies on EXP3 through employing modifies losses.  
\end{itemize}

\begin{figure}[h]
\centering
\begin{minipage}{.48\textwidth}
  \centering
  \includegraphics[width=\linewidth]{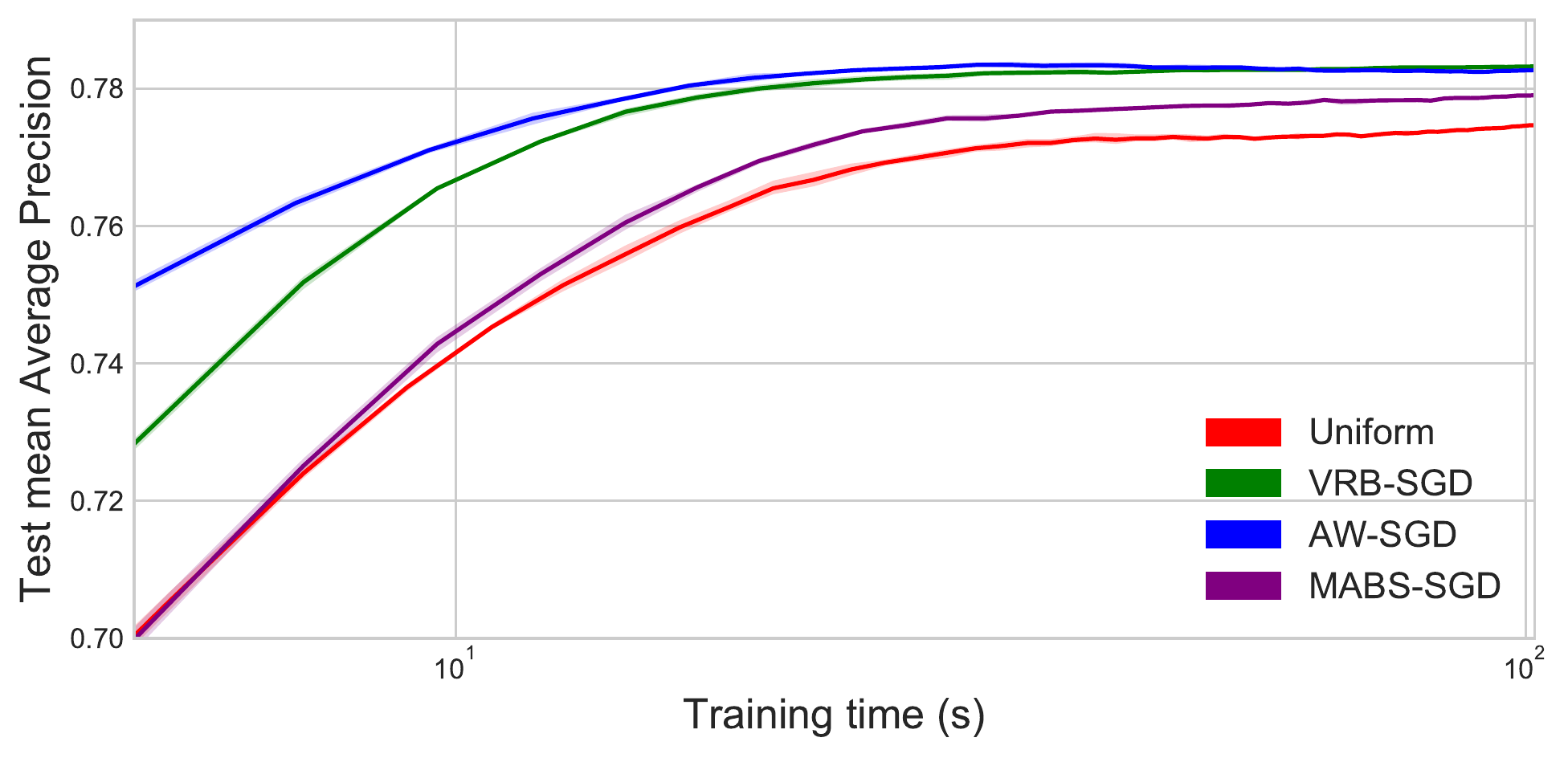}
      \caption{Mean Average Precision scores achieved on the test part of VOC 2007.}
      \label{fig:voc-results}
\end{minipage}%
\quad
\begin{minipage}{.48\textwidth}
  \centering
  \includegraphics[width=\linewidth]{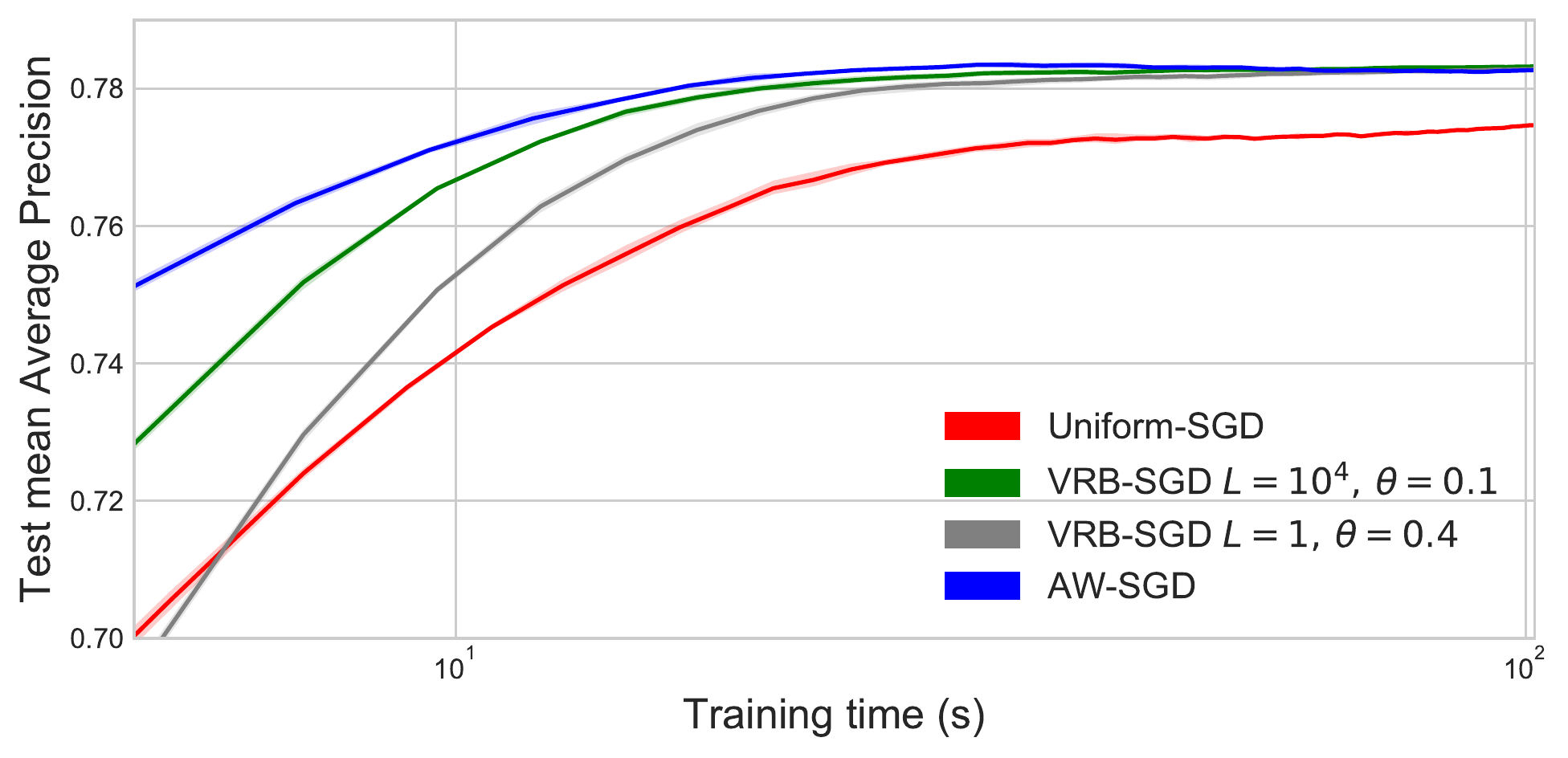}
      \caption{The effect of different hyperparameters on VRB. }
      \label{fig:voc-results-2}
\label{fig:test}
\end{minipage}%
\end{figure}

The hyperparameters of the methods are chosen based on cross-validation on the validation portion of the dataset. The results can be seen in Figure \ref{fig:voc-results}, where the shaded areas represent confidence $95\%$ intervals over 10 runs. The best performing method is AW, but its disadvantage compared to the bandit algorithms is that it requires choosing a family of sampling distributions, which usually incorporates prior knowledge, and calculating the derivative of the log-density. VRB and AW both outperform uniform subsampling with respect to the training time.  VRB performs similarly to AW at convergence, and speeds up training 10 times compared to uniform sampling, by attaining a certain score level 10 times faster. We have also experimented with the variance reduction method of \cite{pmlr-v70-namkoong17a}, but it did not outperform uniform sampling significantly. Since cross-validation is costly, in Figure \ref{fig:voc-results-2} we show the effect of the hyperparameters of our method. More specifically, we compare the performance of VRB with misspecified regularizer $L=1$ to the best $L=10^8$ chosen by cross-validation, and we compensate by using higher mixing coefficient $\theta=0.4$. The fact that only the early-stage performance is affected is a sign of method's robustness against regularizer misspecification.

\subsection{$k$-Means}

In this experiment, we show that in some applications it is beneficial to work with per-sample upper bound estimates $L_i$ instead of a single global bound. As an illustrative example, we choose mini-batch $k$-Means clustering \citep{sculley2010web}. This is a slight deviation from the presented theory, since we sample multiple points for the batch and update the sampler only  once, upon observing the loss for the batch. 

In the case of $k$-Means, the parameters consist of the coordinates of the $k$ centers \linebreak $Q=\{q_1, q_2, \dots, q_k\}$. As the cost function for a point $x_i \in \{x_1, x_2, \dots, x_n \}$ is the  squared Euclidean distance to the closest center, the loss received by VRB  is the norm of the gradient  \linebreak $\min _{q \in Q } 2 \cdot || x_i - q||_2$. This lends itself to a natural estimation of $L_i$:
choose a point $u$ randomly from the dataset and define $L_i=4 \cdot || x_i - u||^2_2$. For this experiment, we set $\theta=0.5$.

We solve mini-batch $k$-Means  for $k=100$ and batch size $b=100$ with uniform sampling and VRB. The initial centers are chosen with $k$-Means++ \citep{arthur2007k} from a random subsample of 1000 points from the training data and they are shared between the methods. We generate 10 different sets of initial centers and run both algorithms 10 times on each set of centers, with different random seeds for the samplers.  We train the algorithm on $80 \%$ of the data, and measure the cost of the $20 \%$ test portion for the following datasets:
\vspace{-2mm}
\begin{itemize}
\setlength\itemsep{0.2em}
  \item \texttt{CSN} \citep{faulkner2011next} --- cellphone accelerometer with 80,000 observations and 17 features,
 \item \texttt{KDD} \citep{kddcup2004} --- data set used for Protein Homology Prediction KDD
    competition  containing 145,751 observations with 74 features,
    \item \texttt{MNIST} \citep{lecun1998gradient} --- 70,000 low resolution images of handwritten
    characters transformed using PCA with whitening and retaining 10 dimensions.  
\end{itemize}

\begin{figure}[h]
      \centering
      \includegraphics[width=\linewidth]{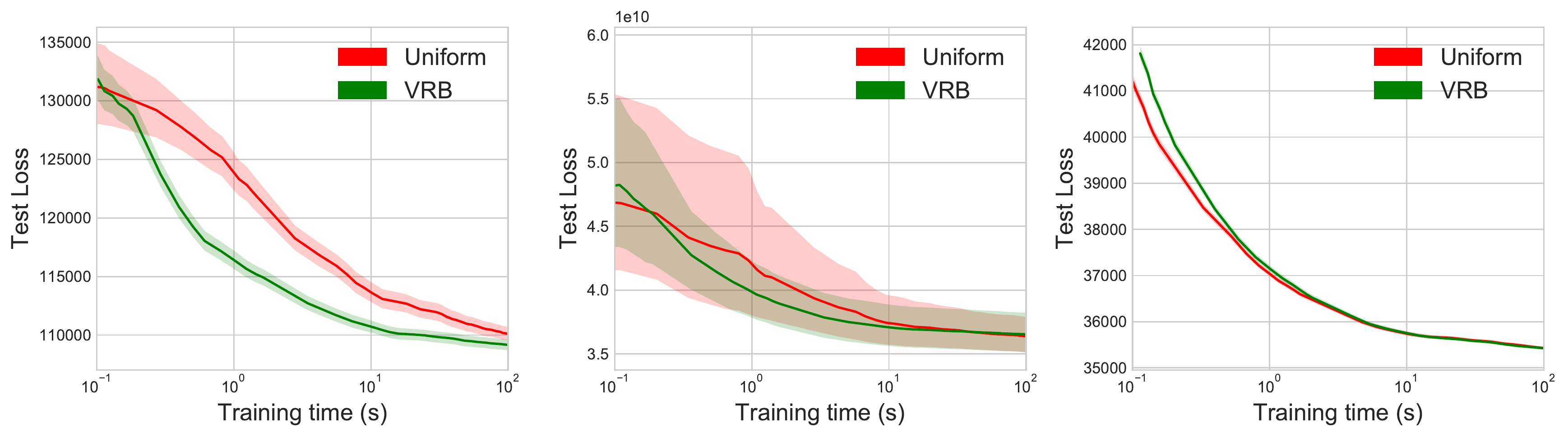}
      \caption{The evolution of the loss of $k$-Means on the test set. The shaded areas represent $95\%$ confidence intervals over 100 runs.}
      \label{fig:kmeans-results}
        \end{figure}
        
The evolution of the cost function on the test set with respect to the elapsed training time is shown in Figure \ref{fig:kmeans-results}. The chosen datasets illustrate three observed behaviors of our algorithm. In the case of \texttt{CSN}, our method significantly outperforms uniform subsampling. In the case of \texttt{KDD}, the advantage of our method can be seen in the reduced variance of the cost over multiple runs, whereas on \texttt{MNIST} we observe no advantage.
This behavior is highly dependent on intrinsic dataset characteristics: for \texttt{MNIST}, we note that the entropy of the best-in-hindsight sampling distribution is close the entropy of the uniform distribution. We have also compared VRB with the bandit algorithms mentioned in the previous section. Since mini-batch $k$-Means converges in 1-2 epochs, these methods with  uniform initialization do not outperform uniform subsampling significantly. Thus, for this setting, careful initialization is necessary, which is naturally supported by our method.

\section{Conclusion and Future Work}

We presented a novel importance sampling technique for variance reduction in an online learning formulation. First, we motivated why regret is a sensible measure of performance in this setting. Despite the bandit feedback and the unbounded costs, we provided an expected regret guarantee of $\tO(n^{1/3}T^{2/3})$, where we reference is the best fixed sampling distribution in hindsight. We confirmed the theoretical findings with empirical validation.

Among the many possible future directions stands the question of the tightness of  the expected regret bound of the algorithm. Another naturally arising idea is theoretical analysis of the method when employed in conjunction with advanced stochastic solvers such as SVRG and SAGA.

\section*{Acknowledgement}
The authors would like  to thank  
Hasheminezhad Seyedrouzbeh  for useful discussions during the course of this work. This research was supported by SNSF grant $407540\_167212$ through the NRP 75 Big Data program.
K.Y.L. is supported by the ETH Zurich Postdoctoral Fellowship and Marie Curie Actions for People COFUND program.

\bibliographystyle{abbrvnat}
\bibliography{bib}

\begin{thebibliography}{35}
\providecommand{\natexlab}[1]{#1}
\providecommand{\url}[1]{\texttt{#1}}
\expandafter\ifx\csname urlstyle\endcsname\relax
  \providecommand{\doi}[1]{doi: #1}\else
  \providecommand{\doi}{doi: \begingroup \urlstyle{rm}\Url}\fi

\bibitem[Abernethy et~al.(2008)Abernethy, Hazan, and Rakhlin]{Abernethy08}
J.~Abernethy, E.~Hazan, and A.~Rakhlin.
\newblock Competing in the dark: An efficient algorithm for bandit linear
  optimization.
\newblock In \emph{COLT}, pages 263--274, 2008.

\bibitem[Alain et~al.(2015)Alain, Lamb, Sankar, Courville, and
  Bengio]{alain2015variance}
G.~Alain, A.~Lamb, C.~Sankar, A.~Courville, and Y.~Bengio.
\newblock Variance reduction in sgd by distributed importance sampling.
\newblock \emph{arXiv preprint arXiv:1511.06481}, 2015.

\bibitem[Allen-Zhu et~al.(2016)Allen-Zhu, Qu, Richt{\'a}rik, and
  Yuan]{allen2016even}
Z.~Allen-Zhu, Z.~Qu, P.~Richt{\'a}rik, and Y.~Yuan.
\newblock Even faster accelerated coordinate descent using non-uniform
  sampling.
\newblock In \emph{International Conference on Machine Learning}, pages
  1110--1119, 2016.

\bibitem[Arthur and Vassilvitskii(2007)]{arthur2007k}
D.~Arthur and S.~Vassilvitskii.
\newblock k-means++: The advantages of careful seeding.
\newblock In \emph{Proceedings of the eighteenth annual ACM-SIAM symposium on
  Discrete algorithms}, pages 1027--1035. Society for Industrial and Applied
  Mathematics, 2007.

\bibitem[Auer et~al.(2002)Auer, Cesa-Bianchi, Freund, and
  Schapire]{auer2002nonstochastic}
P.~Auer, N.~Cesa-Bianchi, Y.~Freund, and R.~E. Schapire.
\newblock The nonstochastic multiarmed bandit problem.
\newblock \emph{SIAM journal on computing}, 32\penalty0 (1):\penalty0 48--77,
  2002.

\bibitem[Bottou and Bengio(1995)]{bottou1995convergence}
L.~Bottou and Y.~Bengio.
\newblock Convergence properties of the k-means algorithms.
\newblock In \emph{Advances in neural information processing systems}, pages
  585--592, 1995.

\bibitem[Bouchard et~al.(2015)Bouchard, Trouillon, Perez, and
  Gaidon]{bouchard2015online}
G.~Bouchard, T.~Trouillon, J.~Perez, and A.~Gaidon.
\newblock Online learning to sample.
\newblock \emph{arXiv preprint arXiv:1506.09016}, 2015.

\bibitem[Cesa-Bianchi et~al.(2004)Cesa-Bianchi, Conconi, and
  Gentile]{cesa2004generalization}
N.~Cesa-Bianchi, A.~Conconi, and C.~Gentile.
\newblock On the generalization ability of on-line learning algorithms.
\newblock \emph{IEEE Transactions on Information Theory}, 50\penalty0
  (9):\penalty0 2050--2057, 2004.

\bibitem[Chawla et~al.(2002)Chawla, Bowyer, Hall, and
  Kegelmeyer]{chawla2002smote}
N.~V. Chawla, K.~W. Bowyer, L.~O. Hall, and W.~P. Kegelmeyer.
\newblock Smote: synthetic minority over-sampling technique.
\newblock \emph{Journal of artificial intelligence research}, 16:\penalty0
  321--357, 2002.

\bibitem[Csiba and Richt{\'a}rik(2016)]{csiba2016importance}
D.~Csiba and P.~Richt{\'a}rik.
\newblock Importance sampling for minibatches.
\newblock \emph{arXiv preprint arXiv:1602.02283}, 2016.

\bibitem[Defazio et~al.(2014)Defazio, Bach, and
  Lacoste-Julien]{defazio2014saga}
A.~Defazio, F.~Bach, and S.~Lacoste-Julien.
\newblock Saga: A fast incremental gradient method with support for
  non-strongly convex composite objectives.
\newblock In \emph{Advances in Neural Information Processing Systems}, pages
  1646--1654, 2014.

\bibitem[Duchi et~al.(2011)Duchi, Hazan, and Singer]{duchi2011adaptive}
J.~Duchi, E.~Hazan, and Y.~Singer.
\newblock Adaptive subgradient methods for online learning and stochastic
  optimization.
\newblock \emph{Journal of Machine Learning Research}, 12\penalty0
  (Jul):\penalty0 2121--2159, 2011.

\bibitem[Everingham et~al.(2010)Everingham, Van~Gool, Williams, Winn, and
  Zisserman]{everingham2010pascal}
M.~Everingham, L.~Van~Gool, C.~K. Williams, J.~Winn, and A.~Zisserman.
\newblock The pascal visual object classes (voc) challenge.
\newblock \emph{International journal of computer vision}, 88\penalty0
  (2):\penalty0 303--338, 2010.

\bibitem[Faulkner et~al.(2011)Faulkner, Olson, Chandy, Krause, Chandy, and
  Krause]{faulkner2011next}
M.~Faulkner, M.~Olson, R.~Chandy, J.~Krause, K.~M. Chandy, and A.~Krause.
\newblock The next big one: Detecting earthquakes and other rare events from
  community-based sensors.
\newblock In \emph{Information Processing in Sensor Networks (IPSN), 2011 10th
  International Conference on}, pages 13--24. IEEE, 2011.

\bibitem[Freedman(1975)]{freedman1975tail}
D.~A. Freedman.
\newblock On tail probabilities for martingales.
\newblock \emph{the Annals of Probability}, pages 100--118, 1975.

\bibitem[Hazan(2011)]{Hazan09}
E.~Hazan.
\newblock A survey: The convex optimization approach to regret minimization.
\newblock \emph{Optimization for machine learning}, pages 287--302, 2011.

\bibitem[Johnson and Zhang(2013)]{johnson2013accelerating}
R.~Johnson and T.~Zhang.
\newblock Accelerating stochastic gradient descent using predictive variance
  reduction.
\newblock In \emph{Advances in neural information processing systems}, pages
  315--323, 2013.

\bibitem[Kakade and Tewari(2009)]{kakade2009generalization}
S.~M. Kakade and A.~Tewari.
\newblock On the generalization ability of online strongly convex programming
  algorithms.
\newblock In \emph{Advances in Neural Information Processing Systems}, pages
  801--808, 2009.

\bibitem[Kalai and Vempala(2005)]{kalai2005efficient}
A.~Kalai and S.~Vempala.
\newblock Efficient algorithms for online decision problems.
\newblock \emph{Journal of Computer and System Sciences}, 71\penalty0
  (3):\penalty0 291--307, 2005.

\bibitem[KDD Cup 2004()]{kddcup2004}
KDD Cup 2004.
\newblock {KDD Cup 2004. Protein Homology Dataset.}
\newblock \url{http://osmot.cs.cornell.edu/kddcup/}, 2004.
\newblock Accessed: 10.11.2016.

\bibitem[LeCun et~al.(1998)LeCun, Bottou, Bengio, and
  Haffner]{lecun1998gradient}
Y.~LeCun, L.~Bottou, Y.~Bengio, and P.~Haffner.
\newblock {Gradient-based learning applied to document recognition}.
\newblock \emph{Proceedings of the IEEE}, 86\penalty0 (11):\penalty0
  2278--2324, 1998.

\bibitem[McMahan and Streeter(2010)]{mcmahan2010adaptive}
H.~B. McMahan and M.~Streeter.
\newblock Adaptive bound optimization for online convex optimization.
\newblock \emph{COLT 2010}, page 244, 2010.

\bibitem[Namkoong et~al.(2017)Namkoong, Sinha, Yadlowsky, and
  Duchi]{pmlr-v70-namkoong17a}
H.~Namkoong, A.~Sinha, S.~Yadlowsky, and J.~C. Duchi.
\newblock Adaptive sampling probabilities for non-smooth optimization.
\newblock In \emph{Proceedings of the 34th International Conference on Machine
  Learning}, volume~70 of \emph{Proceedings of Machine Learning Research},
  pages 2574--2583, International Convention Centre, Sydney, Australia, 06--11
  Aug 2017. PMLR.

\bibitem[Necoara et~al.(2011)Necoara, Nesterov, and Glineur]{necoara2011random}
I.~Necoara, Y.~Nesterov, and F.~Glineur.
\newblock A random coordinate descent method on large optimization problems
  with linear constraints.
\newblock 2011.

\bibitem[Needell et~al.(2014)Needell, Ward, and Srebro]{needell2014stochastic}
D.~Needell, R.~Ward, and N.~Srebro.
\newblock Stochastic gradient descent, weighted sampling, and the randomized
  kaczmarz algorithm.
\newblock In \emph{Advances in Neural Information Processing Systems}, pages
  1017--1025, 2014.

\bibitem[Nesterov(2012)]{nesterov2012efficiency}
Y.~Nesterov.
\newblock Efficiency of coordinate descent methods on huge-scale optimization
  problems.
\newblock \emph{SIAM Journal on Optimization}, 22\penalty0 (2):\penalty0
  341--362, 2012.

\bibitem[Perekrestenko et~al.(2017)Perekrestenko, Cevher, and
  Jaggi]{perekrestenko2017faster}
D.~Perekrestenko, V.~Cevher, and M.~Jaggi.
\newblock Faster coordinate descent via adaptive importance sampling.
\newblock In \emph{Proceedings of the 20th International Conference on
  Artificial Intelligence and Statistics}, volume~54. PMLR, 2017.

\bibitem[{Salehi} et~al.(2017){Salehi}, {Celis}, and {Thiran}]{salehi2017}
F.~{Salehi}, L.~E. {Celis}, and P.~{Thiran}.
\newblock {Stochastic Optimization with Bandit Sampling}.
\newblock \emph{ArXiv e-prints}, Aug. 2017.

\bibitem[Salehi et~al.(2017)Salehi, Thiran, and Celis]{salehi2017stochastic}
F.~Salehi, P.~Thiran, and L.~E. Celis.
\newblock Stochastic dual coordinate descent with bandit sampling.
\newblock \emph{arXiv preprint arXiv:1712.03010}, 2017.

\bibitem[Sculley(2010)]{sculley2010web}
D.~Sculley.
\newblock Web-scale k-means clustering.
\newblock In \emph{Proceedings of the 19th international conference on World
  wide web}, pages 1177--1178. ACM, 2010.

\bibitem[Shalev-Shwartz et~al.(2012)]{shalev2012online}
S.~Shalev-Shwartz et~al.
\newblock Online learning and online convex optimization.
\newblock \emph{Foundations and Trends{\textregistered} in Machine Learning},
  4\penalty0 (2):\penalty0 107--194, 2012.

\bibitem[Shrivastava et~al.(2016)Shrivastava, Gupta, and
  Girshick]{shrivastava2016training}
A.~Shrivastava, A.~Gupta, and R.~Girshick.
\newblock Training region-based object detectors with online hard example
  mining.
\newblock In \emph{Proceedings of the IEEE Conference on Computer Vision and
  Pattern Recognition}, pages 761--769, 2016.

\bibitem[Simonyan and Zisserman(2015)]{simonyan2014very}
K.~Simonyan and A.~Zisserman.
\newblock Very deep convolutional networks for large-scale image recognition.
\newblock \emph{ICLR}, 2015.

\bibitem[Stich et~al.(2017)Stich, Raj, and Jaggi]{NIPS2017_7025}
S.~U. Stich, A.~Raj, and M.~Jaggi.
\newblock Safe adaptive importance sampling.
\newblock In \emph{Advances in Neural Information Processing Systems 30}, pages
  4384--4394. Curran Associates, Inc., 2017.

\bibitem[Zhao and Zhang(2015)]{zhao2015stochastic}
P.~Zhao and T.~Zhang.
\newblock Stochastic optimization with importance sampling for regularized loss
  minimization.
\newblock In \emph{Proceedings of the 32nd International Conference on Machine
  Learning (ICML-15)}, pages 1--9, 2015.

\end{thebibliography}


\newpage
\appendix
\section{Cumulative Variance of the Gradients and Quality of Optimization}  \label{appendix:cumulvariance}
The relationship between cumulative second moment of the gradients and quality of optimization has been demonstrated in several works. Since the difference between the second moment and the variance is independent of the sampling distribution $p_t$, the guarantees of our method also translate to guarantees with respect to the cumulative second moments of the gradient estimates. Here we provide two concrete references.

For the following, assume that we would like to minimize a convex objective,
$$
\min_{w\in\W}F(w): = \Expect_{z\sim \D}[f(w;z)]
$$
and we assume that we are able to draw i.i.d. samples from the unknown distribution $\D$.
Thus, given a point $w\in\W$ we are able to design an unbiased estimate for $\nabla F(w)$ by sampling $z\sim\D$ and taking $g: = \nabla f(w;z)$ (clearly,
$
\expval{g\vert w} = \nabla F(w)
$).
Now assume a gradient-based update rule, i.e.,
\begin{align}\label{eq:GDupdate}
w_{t+1} = \Pi_{\W}(w_t -\eta_t g_t),\quad \text{where}~~~\expval{g_t\vert w_t} = \nabla F(w_t)
\end{align}
and $\Pi_\W(u): = \argmin_{w\in\W}\|u-w\|$.
Next we  show that for two very popular gradient based-methods --- AdaGrad and SGD for strongly-convex functions, the performance is directly related to the cumulative second moment of the gradient estimates, $\sumtt \Expect\|g_t\|^2$. The latter is exactly the objective of our online variance reduction method.

The AdaGrad algorithm employs the same rule as in Eq.~\eqref{eq:GDupdate} using
$\eta_t = D/\sqrt{2\sum_{\tau=1}^t \|g_t\|^2}$. The next theorem substantiates its guarantees.
\begin{theorem}[\cite{duchi2011adaptive}] Assume that the diameter of $\W$ is bounded by $D$. Then:

\begin{equation*}
\expval{F\left(  \frac{1}{T} \sum_{t=1}^T  w_t  \right) } - \min_{w\in\W}F(w) \leq \frac{2D}{ T}\sqrt{ \sumtt \Expect\|g_t\|^2}
\end{equation*}
\end{theorem}

The SGD algorithm for $\mu$-strongly-convex objectives employs the same rule as in Eq.~\eqref{eq:GDupdate} using
$\eta_t = \frac{2}{\mu t}$. The next theorem substantiates its guarantees.

\begin{theorem}[\cite{salehi2017}] Assume that $F$ is $\mu$-strongly convex, then:
\begin{equation*}
\expval{F\left(  \frac{2}{T(T+1)} \sum_{t=1}^T t\cdot w_t  \right) } - \min_{w\in\W}F(w) \leq \frac{2}{\mu T(T+1)} \sumtt \Expect\|g_t\|^2
\end{equation*}
\end{theorem}

\section{Proof of Lemma~\ref{lem:MeaningfulRegret}}
\begin{proof}
Denote $\ell^2_{1:t}(i) = \sum_{\tau=1}^t \ell_\tau^2(i)$.
Next, we bound the  cumulative  loss per point $i\in[n]$,
 \begin{align} \label{eq:SumeLL}
  \ell_{1:T}^2(i)= \sum_{t=1}^T {\ell_t^2(i)} 
& =  \sum_{t=1}^T {\left(\ell_*(i) + \left(\ell_t(i)-\ell_*(i)\right) \right)^2}   \nonumber\\
&\leq T\cdot \ell_{*}^2(i) + 2\ell_{*}(i)\sum_{t=1}^T|\ell_t(i) - \ell_{*}(i)| + \sum_{t=1}^T(\ell_t(i) - \ell_{*}(i))^2 \nonumber\\
&\leq T\cdot \ell_{*}^2(i) + 2\ell_{*}(i)\sqrt{T\cdot V_T(i)} +V_T(i)  \nonumber\\
&=T\left(\ell_{*}(i) +   \sqrt{\frac{V_T(i)}{T}} \right)^2
\end{align}
where the second line uses $\ell_*(i)\geq 0$ and the third line uses the definition of $V_T(i)$ together with the inequality
$\|u\|_1\leq \sqrt{T}\|u\|_2,\; \forall u\in \reals^T$.

We require the following lemma:
\begin{lemma}\label{lem:OptValue}
Let $a_1,\ldots,a_n\geq0$. Then the following holds,
$$
\min_{p\in\Delta}\sumin\frac{a_i}{p(i)} =\left( \sumin\sqrt{a_i}\right)^2~.
$$
\end{lemma}
The proof of the lemma is analogous to the proof of Lemma \ref{lemma:ftrl-sol}, which is given in the next section. Notice that according to this lemma and using the non-negativity of losses we have,
\begin{align} \label{eq:OptimalPerpoint}
\frac{1}{n^2}\min_{p\in\Delta}\sumin \frac{\ell_t^2(i)}{p(i)} =\left(\frac{1}{n} \sumin \ell_t(i) \right)^2:= L^2(w_t)~.
\end{align}
We are now ready to bound the value of best fixed point in hindsight,
\begin{align*}
\min_{p} \frac{1}{n^2}\sum_{t=1}^T \sumin \frac{\ell_t^2(i)}{p(i)} 
& = 
\min_{p} \frac{1}{n^2}\sumin \frac{\ell_{1:T}^2(i)}{p(i)} \\
& = 
 \frac{1}{n^2}\left( \sumin  \sqrt{\ell^2_{1:t}(i)} \right)^2 \\
& = 
T \left( \frac{1}{n}\sumin \ell_{*}(i) +\frac{1}{n}\sumin  \sqrt{\frac{V_T(i)}{T}}  \right)^2 \\
 &=
 T \cdot L_*^2 + 2\sqrt{T}L_* \cdot \frac{1}{n}\sumin  \sqrt{{V_T(i)}} + \left(\frac{1}{n}\sumin  \sqrt{{V_T(i)}} \right)^2~,
\end{align*}
where in the second line we use Lemma~\ref{lem:OptValue}, and the third line uses Eq.~\eqref{eq:SumeLL}.

We are now left to prove that $T \cdot L_*^2\leq \sumtt \frac{1}{n^2}\min_{p\in\Delta}\sumin {\ell_t^2(i)}/{p(i)} $. Indeed, \begin{align*}
  L_*^2 
  &\leq
 \left( \frac{1}{T}\sumtt L(w_t)\right)^2 \\
 &\leq
 \frac{1}{T}\sumtt L^2(w_t) \\
 &=
 \frac{1}{T}\sumtt  \frac{1}{n^2}\min_{p\in\Delta}\sumin {\ell_t^2(i)}/{p(i)}~.
\end{align*}
where the first line uses the asuumption about the average optimality of $L_*$, the second line uses Jensen's inequality, and the last line uses Eq.~\eqref{eq:OptimalPerpoint}. 
This concludes the proof.
\end{proof}

\section{Proofs for the Full Information Setting }
\subsection{ Proof of Lemma \ref{lemma:ftrl-sol}}
\begin{proofarg}{}
We formulate the Lagrangian of the optimization problem in Equation \eqref{eq:ftrl-def}:
\begin{equation*}
\begin{aligned}
& \underset{p}{\text{minimize}}
& &  \sum_{i=1}^n \frac{\ell_{1:t-1}^2(i)}{p(i)}  + \gamma \sum_{i=1}^n \frac{1}{p(i)} \\
& \text{subject to}
& & \sum_{i=1}^n p(i)=1\\ 
& & &p(i) \geq 0, \; i = 1, \ldots, n
\end{aligned}
\end{equation*}
and get:
\begin{equation*}
\mathcal{L}(p, \lambda) = \sum_{i=1}^n \frac{\ell_{1:t-1}^2(i)}{p(i)} + \gamma \sum_{i=1}^n \frac{1}{p(i)} + \alpha \cdot \left(\sum_{i=1}^n p(i) - 1 \right) -\sum_{i=1}^n \beta_i \cdot p(i)
\end{equation*}
From setting $\frac{\partial \mathcal{L}(p, \lambda)}{\partial p(i)}=0 $ we have:
\begin{equation}
p(i) = \frac{\sqrt{\ell_{1:t-1}^2(i)+\gamma}}{\sqrt{\alpha - \beta_i}}
\end{equation}
Note that setting $p(i)=0$ implies an objective value of infinity due to the regularizer. Thus, at the optimum $p(i)>0,\; \forall i\in[n]$;
 which in turn implies that  $\beta_i=0,\; \forall i\in[n]$ (due to complementary slackness).
Combining this with $\sum_{i=1}^n p(i)=1$, we get $\sqrt{\alpha} = \sum_{i=1}^n \sqrt{\ell_{1:t-1}^2(i)+\gamma} $ which gives:
\begin{equation}
p(i) = \frac{\sqrt{\ell_{1:t-1}^2(i)+\gamma}}{ \sum_{j=1}^n \sqrt{\ell_{1:t-1}^2(j)+\gamma} }
\end{equation}
Since the minimization problem is convex for $p\in \Delta$, we obtained a global minimum.
\end{proofarg}

\subsection{ Proof of Lemma \ref{lemma:ub-a}}
\begin{proofarg}{}
The regret of FTRL may be related to the stability of the online decision sequence as shown in
 the following lemma due to \cite{kalai2005efficient} (proof 
 can be found in \cite{Hazan09}  or in \cite{shalev2012online}):
\begin{lemma} \label{Lemma:FTL-BTL_appendix}
Let $\K$ be a convex set and  $\R:\K\mapsto\reals$ be a regularizer. Given a sequence of cost functions $\{f_t \}_{t\in[T]}$ defined over $\K$, then setting $p_t = \argmin_{p\in\Delta}\sum_{\tau=1}^{t-1}f_{\tau}(p)+\R(p)$ 
ensures
\begin{align*}
\sum_{t=1}^T f_t(p_t)-\sum_{t=1}^T f_t(p)\leq \sum_{t=1}^T\left( f_t(p_t) -f_t(p_{t+1}) \right) +(\R(p)-\R(p_1)), \quad \forall p\in\K.
\end{align*}
\end{lemma}
Notice that our regularizer $\R(p) =  L \sum_{i=1}^n {1}/{p(i)}$ is non-negative and bounded by $nL/\pmin$ over $\Delta'$. 
Thus, applying the above lemma to the FTRL rule of Eq.~\eqref{eq:ftrl-def} implies that $\forall p\in\Delta'$,
\begin{equation} \label{eq:ftrl-regret}
\sumtt f_t(p_t)   - \sumtt f_t(p)  
\leq  \sumtt \left( f_t(p_t)- f_t(p_{t+1}) \right) +  \frac{nL}{\pmin}~.
\end{equation}
We are left to bound the remaining term. Let us first recall the closed from solution for the $p_t$'s as stated in Lemma~\ref{lemma:ftrl-sol},
$$
p_t(i) =\frac{\sqrt{\ell_{1:t-1}^2(i) + L}}{c_t}~,
$$
where $c_t = \sumin \sqrt{\ell_{1:{t-1}}^2(i) + L} $ is the normalization factor. Noticing that $\{c_t\}_{t\in[T]}$ is a non-decreasing sequence
we, are now ready to bound the remaining term,
\begin{align*}
\sumtt \left(   f_t(p_t) -  f_t(p_{t+1})  \right)  &= \sum_{t=1}^T \sumin \ell_t^2(i) \cdot \left( \frac{c_t }{\sqrt{\ell^2_{1:t-1}(i) +L}  }-\frac{c_{t+1} }{\sqrt{\ell^2_{1:t}(i) +L}  }   \right)\\
&\leq \sum_{t=1}^T \sumin \ell_t^2(i) \cdot \left( \frac{c_t }{\sqrt{\ell^2_{1:t-1}(i) +L}  }-\frac{c_t }{\sqrt{\ell^2_{1:t}(i) +L}  }   \right)\\
&=\sum_{t=1}^T \sumin \frac{\ell_t^2(i)\cdot c_t}{\sqrt{\ell^2_{1:t}(i) +L} } \cdot \left( \sqrt{1+ \frac{\ell_t^2(i)}{\ell^2_{1:t-1}(i) +L} } -1  \right)\\
&\leq 
\frac{c_T}{2}\sum_{t=1}^T \sumin  \frac{  \ell_t^4(i)}{\sqrt{\ell_{1:t}^2(i)+L} \cdot \left( \ell^2_{1:t-1}(i) +L \right)}
\end{align*}
where in the first inequality we used the fact that $c_t \leq c_{t+1}$ and in the last inequality we relied on the fact that $\sqrt{1+x} \leq 1+\frac{x}{2}$ for all $x \geq 0$. Furthermore, we observe that $\sqrt{\ell_{1:t}^2(i)+L}  \geq \sqrt{\ell_{1:t}^2(i)} $ and $\ell^2_{1:t-1}(i) +L \geq \ell^2_{1:t}(i)$ in order to get:
\begin{equation*}
\sumtt \left(   f_t(p_t) -  f_t(p_{t+1})  \right) \leq \frac{c_T}{2} \cdot \sum_{t=1}^T \sumin  \frac{\ell_t^4(i)}{\left(\ell_{1:t}^2(i)\right)^{3/2}} =  \sqrt{L} \cdot \frac{c_T}{2}  \cdot \sum_{i=1}^n \sum_{t=1}^T\frac{ \frac{\ell_t^4(i)}{L^2}  }{\left( \frac{\ell_{1:t}^2(i)}{L}  \right)^{3/2}} 
\end{equation*}
For a fixed index $i$,  denote $a_t: = \ell_t(i) / \sqrt{L}$  and note that $a_t\in [0, 1],\; \forall t \in [T]$. The innermost sum can be therefore written as $\sum_{t=1}^T \frac{a_t^4}{(a^2_{1:t})^{3/2}}$, which is upper bounded by $44$ as stated in lemma below.
\begin{lemma}\label{lem:SumConst}
For any sequence of  numbers $a_1,\ldots, a_T \in[0,1]$ the following holds:
\begin{equation*}
\sum_{t=1}^T \frac{a_t^4}{(a^2_{1:t})^{3/2}} \leq 44~.
\end{equation*}
\end{lemma}
The proof of the lemma is provided in section \ref{subsec:proof-lem-SumCost}. As a consequence,
\begin{align}\label{eq:ftrl-ub1}
\sumtt \left(   f_t(p_t) -  f_t(p_{t+1})  \right)
&\leq
  \sqrt{L} \cdot \frac{c_T}{2} \cdot  \sum_{i=1}^n \sum_{t=1}^T\frac{ \frac{\ell_t^4(i)}{L^2}  }{\left( \frac{\ell_{1:t}^2(i)}{L}  \right)^{3/2}} 
  \nonumber\\
  &\leq 
  22 n \sqrt{L} \cdot \sum_{i=1}^n  \sqrt{\ell_{1:T-1}^2(i)+L}~,
\end{align}
where we have used the expression for $c_T$.

We get our final result once we plug Equation \eqref{eq:ftrl-ub1} into Equation \eqref{eq:ftrl-regret} and observe that \linebreak $\sqrt{\ell_{1:T-1}^2(i)+L} \leq  \sqrt{\ell_{1:T}^2(i)}+\sqrt{L}$.
\end{proofarg}

\subsection{Proof of Lemma \ref{lem:SumConst}} \label{subsec:proof-lem-SumCost}
\begin{proof}
Without loss of generality assume that $a_1>0$ (otherwise we can always start the analysis from the first $t$ such that $a_t>0$).
Let us define the following index sets,
\begin{align*}
P_k &= \{ t\in[T]: 4^{k-1} a_1^2 < a^2_{1:t}\leq 4^{k} a_1^2 \}, & \forall k \in \{1, 2, \dots \ceil*{\log_2(1/a_1)}\}
\\
Q_k  &= \{t\in[T]:   k< a^2_{1:t} \leq k+1 \}, & \forall k \in \{1, 2, \dots\}
\end{align*}
The definitions of $P_k$ implies,
\begin{align}
\sum_{t\in P_k} a_t^4 &\leq \left(\sum_{t\in P_k} a_t^2\right)^2 \leq  4^{2k}a_1^4   \label{eq:IndexP}
\end{align}
The definition of  $Q_k$ implies,
\begin{align}
\sum_{t\in Q_k} a_t^4 &\leq \left(\sum_{t\in Q_k} a_t^2\right)^2 \leq 2^2=4     \label{eq:IndexQ}
\end{align}
where the second inequality  uses $\sum_{t\in Q_k}a_t^2 \leq 2$ which follows from the fact that if a set $Q_k$ is non-empty then so is $Q_{k-1}$ (since $a_t\in[0,1]$), and thus,
\begin{align*}
\sum_{t\in Q_k} a_t^2 
&=
 \sum_{t=1}^{T_k} a_t^2 - \sum_{t=1}^{T_{k-1}} a_t^2 \\
 &\leq (k+1) - (k-1) \\
 &=2~.
\end{align*}
where we have defined $T_k :=\max\{t\in[T] :  t\in Q_k\}$.

Using the definitions  of $P_k$ and $Q_k$ together with Equations~\eqref{eq:IndexP},~\eqref{eq:IndexQ}, we get,
\begin{align*}
\sum_{t=1}^T \frac{a_t^4}{(a^2_{1:t})^{3/2}} & \leq a_1 + \sum_{k=1}^{\ceil{\log_2(1/a_1)}} \sum_{t \in P_k} \frac{a_t^4}{(a^2_{1:t})^{3/2}} +\sum_{k=1}^{\infty} \sum_{t \in Q_k} \frac{a_t^4}{(a^2_{1:t})^{3/2}} \\
&\leq
 a_1 + \sum_{k=1}^{\ceil{\log_2(1/a_1)}} \frac{\sum_{t\in P_k} a_t^4}{4^{3(k-1)/2} a_1^3} 
+ \sum_{k=1}^{\infty}  \frac{\sum_{t\in Q_k} a_t^4}{k^{3/2}} \\
&\leq
 a_1 + \sum_{k=1}^{\ceil{\log_2(1/a_1)}} \frac{4^{2k}a_1^4}{4^{3(k-1)/2} a_1^3} 
+ \sum_{k=1}^{\infty}  \frac{4}{k^{3/2}} \\
&\leq
a_1 \cdot  \left( 1 +\sum_{k=1}^{\ceil{\log_2(1/a_1)}} \frac{ 4^{2k}}{4^{3(k-1)/2} } \right)
+  \sum_{k=1}^{\infty} \frac{4}{k^{3/2}} \\
&\leq
a_1 \cdot   \sum_{k=0}^{\ceil{\log_2(1/a_1)}} 2^{k+3} 
+ 4 \cdot \sum_{k=1}^{\infty} \frac{1}{k^{3/2}} \\
&\leq
16a_1 \cdot 2^{\ceil{\log_2(1/a_1)}} 
+ 4 + 4 \cdot \sum_{k=2}^{\infty} \frac{1}{k^{3/2}} \\
&\leq
 36 + 4 \cdot \sum_{k=2}^{\infty} \frac{1}{k^{3/2}} \\
&\leq
36
+4 \cdot \int_{x=1}^\infty \frac{1}{x^{3/2}}dx \\
&\leq
44
\end{align*}
which concludes the proof.
\end{proof}

\subsection{Proof of Lemma \ref{lemma:ub-b}}
\begin{proofarg}{}
We first look at the loss of the best distribution in hindsight:
\begin{equation*}
\begin{aligned}
& \underset{p}{\text{minimize}}
& & \sum_{t=1}^{T} \sum_{i=1}^n \frac{\ell_t^2(i)}{p(i)} \\
& \text{subject to}
& & \sum_{i=1}^n p(i)=1\\ 
& & &p(i) \geq 0, \; i = 1, \ldots, n.
\end{aligned}
\end{equation*}
Analogous reasoning to the proof of Lemma \ref{lemma:ftrl-sol} we get $p(i) \propto \sqrt{\ell^2_{1:T}(i)}$ and as a consequence, the loss of the best distribution in hindsight over the unrestricted simplex is:
\begin{equation} \min_{p \in \Delta} \sum_{t=1}^{T} \sum_{i=1}^n \frac{\ell_t^2(i)}{p(i)} \label{eq:ftrl-opt}
 = \left(\sumin \sqrt{\ell^2_{1:T}(i)} \right)^2
\end{equation}

The next step is to solve the optimization problem over the restricted simplex $\Delta'$:
\begin{equation*}
\begin{aligned}
& \underset{p}{\text{minimize}}
& & \sum_{t=1}^{T} \sum_{i=1}^n \frac{\ell_t^2(i)}{p(i)} \\
& \text{subject to}
& & \sum_{i=1}^n p(i)=1\\ 
& & &p(i) \geq \pmin, \; i = 1, \ldots, n.
\end{aligned}
\end{equation*}

We start our proof similarly to the proof of Proposition 5 of \cite{pmlr-v70-namkoong17a}.  First, we formulate the Lagrangian:
\begin{equation}
\mathcal{L}(p, \lambda, \theta) = \sumin \frac{\ell_{1:T}^2(i) }{p(i)} + \alpha \cdot \left( \sumin p(i) - 1 \right) - \sumin \beta_i \cdot ( p(i) - \pmin)
\end{equation}
Setting $\frac{\partial \mathcal{L} }{ \partial p(i)} =0 $ and using complementary slackness we get:
\begin{equation} \label{eq:pi-def}
p(i)= \frac{\sqrt{\ell_{1:T}^2(i)}}{\sqrt{\alpha-\beta_i}} = 
\left\{\begin{matrix}
 \frac{\sqrt{\ell_{1:T}^2(i)}}{\sqrt{\alpha}} &  \textup{if } \sqrt{\ell_{1:T}^2(i)}>\sqrt{\alpha} \cdot \pmin \\ 
\pmin & \textup{else}
\end{matrix}\right.
\end{equation}

Next we determine the value of $\alpha$. Denoting $I=\{ i \mid \sqrt{\ell_{1:T}^2(i)}>\sqrt{\alpha} \cdot \pmin\}$,  and using   $\sumin p(i)=1$ implies,
\begin{equation*}
\sumin p(i) =\sum_{i\in I} p(i) + \sum_{i \in I^C} p(i) = \frac{1}{\sqrt{\alpha}} \sum_{i\in I}\sqrt{\ell_{1:T}^2(i)} + (n-|I|) \cdot \pmin = 1
\end{equation*}
From this we get,
\begin{equation} \label{eq:lambda-form}
\sqrt{\alpha} = \frac{\sum_{i\in I}\sqrt{\ell_{1:T}^2(i)}}{1-(n-|I|) \cdot \pmin}~.
\end{equation}
Now we can plug this into the original problem to get the optimal value:
\begin{align*}
\sumin \frac{\ell_{1:T}^2(i) }{p(i)} &= \sum_{i\in I} \frac{\ell_{1:T}^2(i) }{p(i)} + \sum_{i \in I^C} \frac{\ell_{1:T}^2(i) }{p(i)} \\
&= \sqrt{\alpha} \cdot \left( \sum_{i\in I}\sqrt{\ell_{1:T}^2(i)} \right) + \frac{1}{\pmin}\sum_{i \in I^C} \ell_{1:T}^2(i) && \triangleright \text{Eq. \ref{eq:pi-def}, def. of }p(i) \\
&=\alpha \cdot   (1-(n-|I|) \cdot  \pmin)  + \frac{1}{\pmin}\sum_{i \in I^C} \ell_{1:T}^2(i) &&  \triangleright\text{Eq. \ref{eq:lambda-form}, replacing }\sum_{i\in I}\sqrt{\ell_{1:T}^2(i)}\\
&\leq \alpha \cdot (1-(n-|I|) \cdot \pmin)  + \alpha \cdot \pmin \cdot (n-|I|) &&  \triangleright\text{Eq. \ref{eq:pi-def}, } \ell_{1:T}^2(i) \leq \alpha \pmin^2, \forall i \in I^C \\
&=\alpha \\
&= \frac{\left( \sum_{i\in I}\sqrt{\ell_{1:T}^2(i)} \right)^2}{\left(1-(n-|I|) \pmin\right)^2} &&  \triangleright\text{Eq. \ref{eq:lambda-form}} \\
&\leq \frac{\left( \sum_{i\in I}\sqrt{\ell_{1:T}^2(i)} \right)^2}{\left(1- n \cdot \pmin\right)^2}\\
&\leq \frac{\left( \sumin \sqrt{\ell_{1:T}^2(i)} \right)^2}{\left(1- n \cdot \pmin\right)^2}
\end{align*}

Combining this result with Equation \eqref{eq:ftrl-opt} we obtain,
\begin{equation*}
\min_{p \in \Delta'} \sumin \frac{\ell_{1:T}^2(i) }{p(i)} - \min_{p \in \Delta} \sum_{i=1}^n \frac{\ell_{1:T}^2(i)}{p(i)} \leq \left(  \frac{1}{\left(1- n \cdot \pmin\right)^2} - 1 \right) \cdot \left( \sumin \sqrt{\ell_{1:T}^2(i)} \right)^2
\end{equation*}
Using the fact that $\frac{1}{(1-x)^2}-1 \leq 6x$ for $x \in [0, 1/2]$, with which we are assuming that $\pmin \leq 1/(2n)$, we finally get the claim of the lemma. Note that in the sections following this lemma, all choices of $\pmin$ respect $\pmin \leq 1/(2n)$. 
\end{proofarg}

\section{Proofs for the Pseudo-Regret}
\subsection{Proofs of Theorem \ref{thm:bandit-main}}
\begin{proof}
What remains from the proof sketch is to bound the term $\rA$, which we do here.
Due to the mixing we always have $\tilde{p}_t(i) \geq \theta / n$ for all $t\in [T]$, $i \in [n]$. Moreover $p_t(i) \geq 1 / n$ implies $\tilde{p}_t(i) \geq 1 / n$. Next we upper bound
$
{1}/{\tp_t(i)} - {1}/{p_t(i)}
$. 
If $p_t(i) \leq 1/n$, then the difference is negative, otherwise, 
\begin{align*}
\frac{1}{\tp_t(i)} - \frac{1}{p_t(i)} =\theta \cdot \frac{p_t(i) -\frac{1}{n}}{\tp_t(i) p_t(i)}
< \theta \cdot \frac{p_t(i)}{\tp_t(i) p_t(i)}=\frac{\theta}{\tp_t(i)} \leq n \theta.
\end{align*}
As an immediate consequence we obtain a bound on $\rA$,
\begin{align*}
n^2\cdot \rA:
&=
\expval {\sumtt \tf_t(\tp_t) - \sumtt \tf_t(p_t)  }  \\
&= \expval {\sumtt \sumin  \tell^2_{t}(i) \left( \frac{1}{\tp_t(i)} - \frac{1}{p_t(i)} \right) }  \\
&\leq n  \theta \cdot  \expval { \sumin \tell_{1:T}^2(i) }\\
& \leq n^2  \theta L T~,
\end{align*}
where we used $\ell_t^2(i) \leq L$.
The rest of the proof is completed in the proof sketch.
\end{proof}

\section{Proofs for the Expected Regret} \label{sec:ProofsExpectedRegret}
Throughout the proofs we assume $n \leq T$.
\subsection{Proof of Theorem \ref{thm:bandit-non-oblivious} }
 \begin{proof}
 Using the unbiasedness of the modified costs allows to decompose the regret as follows,
\begin{align} \label{eq:Main} 
n^2\expval{ \regret_T } 
&=  
\expval{   \sumtt f_t(\tp_t)  - \min_{p \in \Delta }  \sumtt f_t(p)}  \nonumber\\
&=
\expval{   \sumtt \tf_t(\tp_t)  - \min_{p \in \Delta }  \sumtt \tf_t(p)}
+\expval{  \min_{p \in \Delta }  \sumtt \tf_t(p)  - \min_{p \in \Delta }  \sumtt f_t(p)}  \nonumber\\
&\leq
n^2\mathcal{O}(Ln^{1/3}T^{2/3}) 
+ 
\expval{  
\underset{\rA}{ \underbrace{
 \left(\sumin \sqrt{\tell_{1:T}^2(i)}  \right)^2 -  \left(\sumin \sqrt{\ell_{1:T}^2(i)}  \right)^2
 }}
 },
\end{align}
where  the last line uses  Equation~\eqref{eq:RegretModifiedCosts}  together with  Jensen's inequality (similarly to the proof of Theorem~\ref{thm:bandit-main}). We have also used the closed form solution for the minimal values of the cumulative true/modified costs, i.e, 
\begin{align*}
\min_{p\in\Delta}\sumtt f_t(p) = \left(\sumin \sqrt{\ell_{1:T}^2(i)}  \right)^2 \textup{   and   }
 \min_{p\in\Delta}\sumtt \tf_t(p) = \left(\sumin \sqrt{\tell_{1:T}^2(i)}  \right)^2,
\end{align*}
the above is established in the proof of Lemma \ref{lemma:ub-b}.

 Thus, in order to establish the theorem, we bound the expectation of  $\rm{(A)}$.
 The high level idea of the proof is to show that for any small enough  $\delta\in[0,1] $ then  w.p. $\geq 1-\delta$ the term $\rm{(A)}$ is bounded by $n^2 \mathcal{O}(n^{1/3}T^{2/3}\log(nT/\delta))$.  Then, by showing that $\rm{(A)}$ is bounded \emph{almost surely},  we are able to  choose a small enough $\delta$ such that  $\expval{(A)} =n^2 \tO(Ln^{1/3}T^{2/3})$. 
 Let us first establish a trivial bound on $\rm{(A)}$,
 \begin{align*}
\rm{(A)} 
&\leq
 \left(\sumin \sqrt{\tell_{1:T}^2(i)}  \right)^2 \\
 &\leq
 \left(\sumin \sqrt{ \frac{\ell_{1:T}^2(i)}{\theta/n}   }  \right)^2 \\
 &=
 Ln^{8/3}T^{4/3}~,
 \end{align*}
 where we used $\ell_{1:T}^2(i) \leq LT$, and $\theta = (n/T)^{1/3}$.
Thus, choosing $1/\delta \geq  Ln^{8/3}T^{4/3}$  ensures that  $\delta \cdot\rm{(A)}\leq 1$ with probability 1.
It now remains to establish a high probability bound for $\rA$.
To do so, we shall bound  the differences $\tell_{1:t}^2(i) -\ell_{1:t}^2(i)$ using  a version of Freedman's concentration inequality \citep{freedman1975tail}.
Later, this will enable us to bound $\rA$. Next we proceed according to these two steps.
 
\paragraph{Step 1: bounding $\tell_{1:t}^2(i) -\ell_{1:t}^2(i)$.\\} 
 Fix $i\in[n]$ and define the following sequence $\{Z_{t,i}: = \tell_t^2(i) - \ell_t^2(i) \}_{t\in[T]}$. Recalling that $\mathbb{E}[\tell_t^2(i) \vert \tp_t,\ell_t] = \ell_t^2(i)$, we have that $\{ Z_{t,i}\}_{t\in[T]}$ is a martingale difference sequence with respect to the filtration $\{\F_{t}\}_{t\in[T]}$ associated with the history of the strategy. Also notice that due to the mixing $|Z_{t,i}| \leq 2| \tell_t^2(i) | \leq 2nL/\theta$. We may bound the conditional variance of the  $Z_{t,i}$ as follows,
 \begin{align}\label{eq:VarExpression}
 \Var(Z_{t,i}\vert \F_{t-1})
 & = 
\expval{ \left( \frac{\ell_t^2(i)}{\tp_t(i)}\mathbbm{1} _{I_t=i} - \ell_t^2(i)   \right)^2\vert \F_{t-1}}  \nonumber\\
 & = 
\expval{ \frac{\ell_t^4(i)}{\tp^2_t(i)}\mathbbm{1} _{I_t=i} -2\frac{\ell_t^4(i)}{\tp_t(i)}\mathbbm{1} _{I_t=i} + \ell_t^4(i)   \vert \F_{t-1}}  \nonumber\\
 &=
 \frac{\ell_t^4(i)}{\tp_t(i)} - \ell_t^4(i) \nonumber\\
 &\leq
 L\frac{\ell_t^2(i)}{\tp_t(i)} 
 \end{align}
 The above characterization of the sequence  $\left\{Z_{t,i} \right\}_{t\in[T]}$ allows us to apply Freedman's concentration inequality that we state below,

\begin{lemma}[Freedman's Inequality \citep{freedman1975tail, kakade2009generalization}]\label{lemma:freedman}
Suppose \linebreak $\{Z_t\}_{t\in[T]}$ is a martingale difference sequence with respect to a filtration $\{\F_t\}_{t\in[T]}$, such that $|Z_t|\leq b$. Define,
$
\Var_t Z_t = \Var\left(Z_t \vert \F_{t-1}\right)
$
and  let $\sigma= \sqrt{\sum_{t=1}^T\Var_t Z_t} $ be the sum of conditional variances of $Z_t$'s. Then for any $\delta\leq 1/e$ and 
$T\geq 3$ we have,
$$
P\left( \sum_{t=1}^TZ_t \geq  \max\left\{2\sigma, 3b\sqrt{\log(1/\delta)} \right\}\sqrt{\log(1/\delta)} 
 \right) \leq  4\delta\log(T)
$$
\end{lemma}

Since  $Z_{1, i}, \dots, Z_{T,i}$ is a martingale difference sequence with $|Z_{t,i}| \leq 2nL/\theta$, we can
  applying the two-sided extension of Lemma \ref{lemma:freedman} to this sequence. Combined with union bound over all \linebreak $i\in[n], t\in[T]$ we have that   $\forall i\in[n], t\in[T]$, then w.p.$\geq 1- 8nT\delta\log(T)$,
 \begin{align}\label{eq:Friedman11}
 | \tell_{1:t}^2(i) - \ell_{1:t}^2(i) |
 &=
  \left| \sum_{\tau=1}^t Z_{\tau,i} \right|  \nonumber\\
   &\leq
   \max \left\{ 2 \sqrt{\sum_{\tau=1}^t \Var(Z_{\tau,i}\vert \F_{\tau-1})}, \frac{6nL}{\theta}\sqrt{\log(1/\delta)} \right\} \sqrt{\log(1/\delta)}  \nonumber\\
 & \leq
 \max \left\{ 2\sigma_i, \frac{6nL}{\theta}\sqrt{\log(1/\delta)} \right\} \sqrt{\log(1/\delta)}. 
 \end{align}
 where we have defined
 $
\sigma_i: = \sqrt{\sum_{t=1}^T \Var(Z_{t,i}\vert \F_{t-1})}
$.  Notice that the last line above uses the fact that $\forall t \in [T]:$ $\sum_{\tau=1}^t \Var(Z_{\tau,i}\vert \F_{\tau-1}) \leq \sigma_i^2$, which holds since the conditional variance is non-negative.

A few remarks are in place before we go on with the proof:
\begin{enumerate}
\item Define $\B$ to be the event that  the bound stated in Equation \eqref{eq:Friedman11} holds. Note that 
$P(\B)  \geq 1- 8nT\delta\log(T)$.
From this point on, all of the statements in the proof  are conditioned on the event $\B$.

\item For ease of notation we shall ignore the $\log(1/\delta)$ terms appearing in Equation \eqref{eq:Friedman11}. Note that these only affect the final guarantees by a factor of $O(\log (nT))$ for the choice of $\delta=1/\textup{poly}(n, T)$.
\item We denote $M_i := \max\{ 2\sigma_i,6nL/\theta \}$. Ignoring  $\log(1/\delta)$ factors, Equation \eqref{eq:Friedman11} can be now restated as follows, $\forall i\in[n], t\in[T]$ w.p.$\geq 1- 8nT\delta\log(T)$,
\begin{align}\label{eq:FriedmanEasy}
 | \tell_{1:t}^2(i) - \ell_{1:t}^2(i) |
 &\leq
M_i
 \end{align}
\end{enumerate}

We are now ready to go on with the proof.
Notice that combining Equations \eqref{eq:FriedmanEasy} and \eqref{eq:VarExpression} provides us with a bound on 
$ |\tell_{1:t}^2(i) - \ell_{1:t}^2(i)|$ which depends on the $\tp_t$'s. The next lemma  provides us with a cleaner bound which gets rid of this dependence. The proof of is provided in Section \ref{subsec:proof-lem-Mbound}.

\begin{lemma}\label{lem:M_bound} 
Conditioning on the event $\B$, the following bound holds,
\begin{align} \label{eq:MboundComplex}
M_i \leq 10n^{\frac{2}{3}}LT^{\frac{1}{3}} + 4\sqrt{L}  \left(\ell_{1:T}^2(i)\right) ^{\frac{1}{4}} \left( \sumin \sqrt{\ell_{1:T}^2(i)}\right)^{\frac{1}{2}},
\end{align}
and also 
\begin{align} \label{eq:MboundSimple}
M_i \leq 14n^{\frac{1}{2}}LT^{\frac{1}{2}}.
\end{align}

\end{lemma}

\paragraph{Step 2: bounding $\rA$.} 
First, we formulate a helper lemma, with its proof provided in Section \ref{subsec:proof-lem-SQRT}.
\begin{lemma}\label{lem:SQRT}
Let $x,a > 0$ then
$$
\sqrt{x+a} - \sqrt{x} \leq \min\{\sqrt{a}, a/\sqrt{x} \}.
$$
\end{lemma}

Equation \eqref{eq:FriedmanEasy} enables us to bound $\rA$ as follows,
\begin{align} \label{eq:BoundC}
\rA
 &=
 \left(\sumin \sqrt{\tell_{1:T}^2(i)}  \right)^2 -  \left(\sumin \sqrt{\ell_{1:T}^2(i)}  \right)^2  \nonumber\\
  &=
 \left(\sumin \sqrt{\ell_{1:T}^2(i)+(\tell_{1:T}^2(i)-\ell_{1:T}^2(i))}  \right)^2 -  \left(\sumin \sqrt{\ell_{1:T}^2(i)}  \right)^2   \nonumber\\
 &\leq
 \left(\sumin \sqrt{\ell_{1:T}^2(i)+M_i}  \right)^2 -  \left(\sumin \sqrt{\ell_{1:T}^2(i)}  \right)^2   \nonumber\\
&=
 \left(\sumin \sqrt{\ell_{1:T}^2(i)+M_i}  +\sumin \sqrt{\ell_{1:T}^2(i)}  \right)
 \cdot
 \left(\sumin \sqrt{\ell_{1:T}^2(i)+M_i}  - \sumin \sqrt{\ell_{1:T}^2(i)}  \right)  \nonumber\\
 &\leq
2\sumin \sqrt{\ell_{1:T}^2(i)+M_i} 
 \cdot
 \left(\sumin \sqrt{\ell_{1:T}^2(i)+M_i}  - \sumin \sqrt{\ell_{1:T}^2(i)}  \right)  \nonumber\\
 &\leq
2\sumin \sqrt{\ell_{1:T}^2(i)} 
 \cdot
  \sumin \min\left\{ \sqrt{M_i} , \frac{M_i}{\sqrt{\ell_{1:T}^2(i)}} \right\}
  +
  2\sumin \sqrt{M_i} 
 \cdot
  \sumin \min\left\{ \sqrt{M_i} , \frac{M_i}{\sqrt{\ell_{1:T}^2(i)}} \right\} \nonumber\\
  &\leq
2\uann{\sumin \sqrt{\ell_{1:T}^2(i)} 
 \cdot
  \sumin \min\left\{ \sqrt{M_i} , \frac{M_i}{\sqrt{\ell_{1:T}^2(i)}} \right\}}{(*)}
  +
  2\uann{\left(\sumin \sqrt{M_i} \right)^2}{(**)},
\end{align}
where the second-to-last  line uses  $\sqrt{a+b} \leq \sqrt{a}+\sqrt{b}$, together with Lemma \ref{lem:SQRT}. 

Let us start with bounding $(**)$,
\begin{align} \label{eq:exp-regret-2nd-term}
 \left( \sumin \sqrt{M_i}  \right)^2
  \leq n^2 \max_{i}M_i 
  &\leq 14 L n^{2 + \frac{1}{2}} T^{\frac{1}{2}} \nonumber\\
  &\leq 14 L n^{2 + \frac{1}{3}} T^{\frac{2}{3}}
\end{align}
where we have used the second part  of Lemma \ref{lem:M_bound}; the second line uses $T\geq n$ leading to 
$(nT)^{1/2}\leq n^{1/3}T^{2/3}$.

The last step of the proof is to bound $(*)$. From Lemma \ref{lem:M_bound}, we have the immediate corollary that,
\begin{equation}\label{eq:MboundMax}
M_i \leq \max \left\{ 16n^{\frac{2}{3}}LT^{\frac{1}{3}}, \, 16\sqrt{L}  \left(\ell_{1:T}^2(i)\right) ^{\frac{1}{4}} \left( \sumin \sqrt{\ell_{1:T}^2(i)}\right)^{\frac{1}{2}} \right\}.
\end{equation}
Denote $i_* = \arg \max_{i \in [n]} \min\left\{ \sqrt{M_i} , \frac{M_i}{\sqrt{\ell_{1:T}^2(i)}} \right\}$.
We divide the remainder of the proof into two cases depending on the argument returned by $\max$ of Eq.~\eqref{eq:MboundMax} for the index $i_*$.
If the $\max$ returns the first argument for $i_*$, i.e. $M_{i_*} = 16n^{\frac{2}{3}}LT^{\frac{1}{3}}$, then
\begin{align}\label{eq:exp-regret-1st-term-a}
 (*) &= \sumin \sqrt{\ell_{1:T}^2(i)} 
 \cdot \sumin \min\left\{ \sqrt{M_i} , \frac{M_{i}}{\sqrt{\ell_{1:T}^2(i)}} \right\} \nonumber\\
 & \leq \sumin \sqrt{\ell_{1:T}^2(i)} 
 \cdot  n \min\left\{ \sqrt{M_{i_*}} , \frac{M_{i_*}}{\sqrt{\ell_{1:T}^2(i_*)}} \right\}  \nonumber\\
 &= 
n \sumin \sqrt{\ell_{1:T}^2(i)} 
 \cdot   \sqrt{M_{i_*}}  \nonumber\\
 &\leq
 n^2\sqrt{LT} \cdot\sqrt{ 16n^{2/3}T^{1/3}} \nonumber\\
 &\leq
 4 n^{2+\frac{1}{3}}L T^\frac{2}{3}.
 \end{align}
In the other case, we have $M_{i_*} = 16\sqrt{L}  \left(\ell_{1:T}^2(i_*)\right) ^{\frac{1}{4}} \left( \sumin \sqrt{\ell_{1:T}^2(i_*)}\right)^{\frac{1}{2}}$. We will need the following lemma, with its proof given Section \ref{subsec:proof-lem-min-pow}:
\begin{lemma}\label{lemma:min-pow}
Fix $w>0$; Let $x\in [0,w]$ and $a,b:\mathbb{R}^+ \rightarrow \mathbb{R}^+ $ functions of $x$,  then
$$
\max_{x\in[0,w]} \min \left\{a(x) \cdot x^{1/8}, b(x) \cdot  x^{-1/4} \right\} \leq \max_{x\in[0,w]} a(x)^{2/3} b(x)^{1/3}.
$$
\end{lemma}
Now we can upper bound $(*)$,
\begin{align}\label{eq:exp-regret-1st-term-b}
 (*) &= \sumin \sqrt{\ell_{1:T}^2(i)} 
 \cdot \sumin \min\left\{ \sqrt{M_i} ,\, \frac{M_i}{\sqrt{\ell_{1:T}^2(i)}} \right\} \nonumber\\
 & \leq n\sumin \sqrt{\ell_{1:T}^2(i)} 
 \cdot   \min\left\{ \sqrt{M_{i^*}} ,\, \frac{M_{i_*}}{\sqrt{\ell_{1:T}^2(i_*)}} \right\}  \nonumber\\
 & \leq n^2 L^\frac{1}{2}T^\frac{1}{2} \cdot   \min\left\{ \sqrt{M_{i_*}} ,\, \frac{M_{i_*}}{\sqrt{\ell_{1:T}^2(i_*)}} \right\} \nonumber\\
 &=  n^2 L^\frac{1}{2}T^\frac{1}{2} \cdot   \min\left\{4L^\frac{1}{4}  \left(\ell_{1:T}^2(i_*)\right) ^{\frac{1}{8}} \left( \sumin \sqrt{\ell_{1:T}^2(i_*)}\right)^{\frac{1}{4}}, \, \frac{16\sqrt{L}  \left( \sumin \sqrt{\ell_{1:T}^2(i_*)}\right)^{\frac{1}{2}}}{\left(\ell_{1:T}^2(i_*)\right) ^{\frac{1}{4}} } \right\}\nonumber\\
 &\overset{(\diamond)}{\leq}  n^2 L^\frac{1}{2}T^\frac{1}{2} \cdot 16 L^{\frac{1}{2}} n^{\frac{1}{3}}  T^{\frac{1}{6}} \nonumber\\
 & \leq 16 n^{2+\frac{1}{3}}L T^\frac{2}{3}~,
 \end{align}
 where for $(\diamond)$ we used Lemma \ref{lemma:min-pow} with $x= {\ell_{1:T}^2(i_*)}$, $a(x)=4L^\frac{1}{4}  \left( \sumin \sqrt{\ell_{1:T}^2(i_*)}\right)^{\frac{1}{4}}$, \linebreak $b(x)=16\sqrt{L}  \left( \sumin \sqrt{\ell_{1:T}^2(i_*)}\right)^{\frac{1}{2}}$, and therefore  $\max_x a(x)^{2/3} b(x)^{1/3} \leq 16 L^{1/2} n^{1/3}  T^{1/6}$.
Combining Equation~\eqref{eq:BoundC} together with Equations~\eqref{eq:exp-regret-2nd-term},\eqref{eq:exp-regret-1st-term-a}, and 
\eqref{eq:exp-regret-1st-term-b}, we may  establish the final bound for $\rA$,
conditioned on the event $\B$:
\begin{equation} \label{eq:final-a-bound}
(A) \leq 64 n^{2+\frac{1}{3}}L T^\frac{2}{3}
\end{equation}

\paragraph{Concluding: } 
Combining Equation \eqref{eq:Main} with \eqref{eq:final-a-bound}  and taking an sufficiently small 
$\delta = 1/\rm{poly}(n,T)$ we have proven the theorem.
\end{proof}

\subsection{ Proof of Lemma~\ref{lem:M_bound}} \label{subsec:proof-lem-Mbound}
\begin{proof}
Recalling that  $M_i= \max\{ 2\sigma_i,6nL/\theta \}$, it is natural to divide the proof into two cases depending on the value of $M_i$.
Since $\theta =(n/T)^{1/3}$, it is immediate to show that the lemma holds for the case where 
$2\sigma_i\leq {6nL}/{\theta}$, since in this case $M_i = {6nL}/{\theta} = 6Ln^{2/3}T^{1/3}$.
The rest of the proof regards the other case where $2\sigma_i>{6nL}/{\theta}$, and therefore 
$M_i = 2\sigma_i$.

\textbf{Step (1): Decomposing $M_i^2$.}
\begin{align} \label{eq:BoundSigma_i}
\frac{1}{4L}M_i^2 
&= 
\frac{1}{L}\sigma_i^2  \nonumber\\
&\leq
 \sum_{t=1}^T \frac{\ell_t^2(i)}{\tp_t(i)}  \nonumber\\
&=
 \sum_{t: \ell_{1:t}^2(i) \leq 2M_i} \frac{\ell_t^2(i)}{\tp_t(i)}
 +
 \sum_{t: \ell_{1:t}^2(i) \geq 2M_i} \ell_t^2(i)\left( \frac{1}{\tp_t(i)} -\frac{1}{p_t(i)} \right)
 +
 \sum_{t: \ell_{1:t}^2(i) \geq 2M_i} \frac{ \ell_t^2(i)}{p_t(i)}
   \nonumber\\
    &\leq
\frac{n}{\theta}\cdot  \sum_{t: \ell_{1:t}^2(i) \leq 2M_i} \ell_t^2(i)
 +
  n  \theta \ell_{1:T}^2(i)
 +
\sum_{t: \ell_{1:t}^2(i) \geq 2M_i} \frac{\ell_t^2(i)}{p_t(i)} \nonumber\\
  &\leq
\frac{2n M_i}{\theta}
 +
 n  \theta L T
 +
  \underset{\textbf{($\star$)}}{\underbrace{\sum_{t: \ell_{1:t}^2(i) \geq 2M_i} \frac{\ell_t^2(i)}{p_t(i)}} },
\end{align}
where in the second line we use the definition of $\sigma_i$ together with the bound of Eq.~\eqref{eq:VarExpression}, implying
 $\sigma_i^2 \leq L \sum_{t=1}^T {\ell_t^2(i)}/{\tp_t(i)}$; in the fourth line we use $\tp_t(i)\geq \frac{\theta}{n}$ (due to mixing), and we also use
$\frac{1}{\tp_t(i)} -\frac{1}{p_t(i)} \leq n\theta$ (see proof of Theorem \ref{thm:bandit-main}). The last line uses $\ell_{1:T}^2(i) \leq LT$.
Next we bound the last term, $\textbf{($\star$)}$.

\textbf{Step (2): Bounding  $\textbf{($\star$)}$.}
We shall first bound $1/p_t(i)$ and later use this in order to bound $\textbf{($\star$)}$. Notice that the following hold $\forall t\in[T]$ such that $ \ell_{1:t}^2(i) \geq 2M_i$, 
\begin{align}
&\quad \tell_{1:t}^2(i)  \geq \ell_{1:t}^2(i) -M_i \geq \frac{1}{2} \ell_{1:t}^2(i)  \label{eq:Losses2M_1}\\
&\quad \tell_{1:t}^2(i)  \leq \ell_{1:t}^2(i) +M_i \leq \frac{3}{2} \ell_{1:t}^2(i), \qquad \forall i\in[n]  \label{eq:Losses2M_2}
\end{align}
where  we have used $ | \tell_{1:t}^2(i) - \ell_{1:t}^2(i) |\leq M_i$ (see Eq~\eqref{eq:FriedmanEasy}), which follows since we condition on the event $\B$.
Combining the above with the definition of $p_t$ (see Lemma \ref{lemma:ftrl-sol}) and denoting $L' := Ln/\theta$, yields,
\begin{align*}
\frac{1}{p_t(i)} 
&=
\frac{\sumin \sqrt{\tell_{1:t-1}^2(i) + L'}}{\sqrt{\tell_{1:t-1}^2(i) + L'}} \\
&\leq
\frac{\sumin \sqrt{\tell_{1:t}^2(i) + L'}}{\sqrt{\tell_{1:t}^2(i) }} \\
&\leq
\sqrt{2}\frac{\sumin \sqrt{\frac{3}{2}\ell_{1:t}^2(i) + L'}}{\sqrt{\ell_{1:t}^2(i) }} \\
&\leq
2\frac{\sumin \sqrt{\ell_{1:T}^2(i) }}{\sqrt{\ell_{1:t}^2(i) }} 
\end{align*}
where in the second line we use $\tell_t^2(i) \leq L'$, in the third we employ Equations~\eqref{eq:Losses2M_1}, \eqref{eq:Losses2M_2};  and  the fourth follows by  noticing $L' = Ln/\theta \leq M_i \leq \ell_{1:t}^2(i)/2$, and also $\ell_{1:t}^2(i)\leq \ell_{1:T}^2(i),\;\forall t\in[T]$.

Using the above inequality we may now bound $\textbf{($\star$)}$,
\begin{align} \label{eq:BoundStar}
\textbf{($\star$)} 
&= 
\sum_{t: \ell_{1:t}^2(i) \geq 2M_i} \frac{\ell_t^2(i)}{p_t(i)}  \nonumber\\
&\leq 
2\left( \sumin \sqrt{\ell_{1:T}^2(i)}\right)  \sum_{t: \ell_{1:t}^2(i) \geq 2M_i} \frac{\ell_t^2(i)}{\sqrt{\ell_{1:t}^2(i)} }  \nonumber\\
&\leq 
2\left( \sumin \sqrt{\ell_{1:T}^2(i)}\right)  \sum_{t=1}^T \frac{\ell_t^2(i)}{\sqrt{\ell_{1:t}^2(i)} }  \nonumber\\
&\leq
4 \left( \sumin \sqrt{\ell_{1:T}^2(i)}\right)  \sqrt{\ell_{1:T}^2(i)} 
\end{align}

where the last inequality uses the following lemma from~\citep{mcmahan2010adaptive}:
\begin{lemma}{\citep{mcmahan2010adaptive}}\label{lem:SqrtSum}
For any non-negative numbers $a_1,\ldots, a_T$ the following holds:
$$\sum_{t=1}^T \frac{a_t}{\sqrt{\sum_{\tau=1}^t a_\tau}} \leq 2\sqrt{\sum_{t=1}^T a_t}$$
\end{lemma}

\textbf{Step (3): Final bound.}
Plugging the bound of Equation \eqref{eq:BoundStar} back into Equation \eqref{eq:BoundSigma_i} implies,

\begin{align*}
\frac{1}{4L}M_i^2 
 &\leq
\frac{2nM_i}{\theta}
 +
  n  \theta L T
 +
 4 \sqrt{\ell_{1:T}^2(i)}\left( \sumin \sqrt{\ell_{1:T}^2(i)}\right) \\
\end{align*}
Denote $a=1/(4L), \, b=2n/\theta, \, c_1=n  \theta L T, \, c_2=4 \sqrt{\ell_{1:T}^2(i)}\left( \sumin \sqrt{\ell_{1:T}^2(i)}\right)$. Then, the above inequality can be reformulated as:
\begin{equation*}
a M_i^2 -b M_i - c_1-c_2 \leq 0.
\end{equation*}
Due to the quadratic formula, the largest $M_i$ that satisfies the inequality above is \linebreak $M_i = \left( b + \sqrt{b^2+4a(c_1+c_2)} \right) / (2a)$. We can get an upper bound on $M_i$ by using \linebreak $\sqrt{b^2+4a(c_1+c_2)} \leq b + 2\sqrt{ac_1}+2\sqrt{ac_2}$ to finally get that
\begin{equation*}
M_i \leq \frac{8nL}{\theta} +2n^{\frac{1}{2}}L \theta ^{\frac{1}{2}} T^{\frac{1}{2}} +  4\sqrt{L}  \left(\ell_{1:T}^2(i)\right) ^{\frac{1}{4}} \left( \sumin \sqrt{\ell_{1:T}^2(i)}\right)^{\frac{1}{2}}.
\end{equation*}
Using  $\theta =(n/T)^{1/3}$ we have proven the first claim of the lemma. For the second claim, we use the upper bound $\ell_{1:T}^2(i) \leq LT$ and note that $n^{\frac{2}{3}} T^{\frac{1}{3}} \leq n^{\frac{1}{2}} T^{\frac{1}{2}}$ (since $T\geq n$).

\end{proof}

\subsection{Proof of Lemma~\ref{lem:SQRT}} \label{subsec:proof-lem-SQRT}
\begin{proof}
\begin{equation*}
(\sqrt{x+a} - \sqrt{x})^2  
= 2x+a - 2\sqrt{x^2+xa}\leq a
\end{equation*}
which proves that $\sqrt{x+a} - \sqrt{x} \leq \sqrt{a}$. On the other hand, we 
have that $\sqrt{x+a} - \sqrt{x} \leq a/\sqrt{x}$, which can be easily seen by rearranging it as $\sqrt{x+a} \leq a/\sqrt{x} + \sqrt{x} $ and taking the square of both side. Combining these two facts we get the results.
\end{proof}

\subsection{Proof of Lemma~ \ref{lemma:min-pow}} \label{subsec:proof-lem-min-pow}
\begin{proof}
Define $F(x): = \min \left\{a(x) \cdot x^{1/8}, b(x) \cdot  x^{-1/4}\right\}$.
Note that in order to establish the lemma it is sufficient to show that the following holds for any $x\geq 0$,
$$F(x)\leq  a(x)^{2/3} b(x)^{1/3}~.$$
To do so, fix $x\geq 0$ and divide into two cases.

\textbf{Case 1:}
If $a(x) x^{1/8}\leq b(x)   x^{-1/4}$ then $x \leq \left({b(x)}/{a(x)}\right)^{8/3}$ implying that  $F(x) = a(x)  x^{1/8} \leq a(x)^{2/3} b(x)^{1/3}$.

\textbf{Case 2:}
If $a(x)  x^{1/8}\geq b(x)   x^{-1/4}$ then $x \geq \left({b(x)}/{a(x)}\right)^{8/3}$ implying that   $F(x) = b(x)  x^{-1/4} \leq a(x)^{2/3} b(x)^{1/3}$.

\end{proof}

\end{document}